\pdfoutput=1
\documentclass[11pt]{article}

\usepackage[final]{acl}

\usepackage{times}
\usepackage{latexsym}

\usepackage[T1]{fontenc}
\usepackage[utf8]{inputenc}

\usepackage{microtype}

\usepackage{inconsolata}

\usepackage{graphicx}
\usepackage{tabularx}
\usepackage{float}
\usepackage{enumitem}
\usepackage{multirow}
\usepackage{soul}
\usepackage{xspace}
\usepackage{pifont}
\usepackage{utfsym}
\usepackage{longtable}
\usepackage{makecell}
\usepackage{booktabs}
\usepackage{amsfonts}
\usepackage{array}

\newcolumntype{C}{>{\centering\arraybackslash}X}
\newcommand{\name}{\texttt{SCOP}\xspace}

\soulregister\name7 
\soulregister\cite7
\soulregister\ref7 
\soulregister\citep7 % 针对\citep命令
\soulregister\citet7 % 针对\citet命令
\soulregister\pageref7 % 针对\pageref命令
\soulregister\name7 
\soulregister\xspace7 
\soulregister\underline7 
\soulregister\ding7 
\soulregister\textcircled7 
\soulregister\normalsize7
\soulregister\small7

\usepackage{enumitem}
\setenumerate[1]{itemsep=1pt,parsep=0pt,topsep=0pt}
\setitemize[1]{itemsep=1pt,parsep=0pt,topsep=0pt,leftmargin=8pt}
\title{\name: Evaluating the Comprehension Process of Large Language Models from a Cognitive View}

\author{
 \textbf{Yongjie Xiao\textsuperscript{1, 2}},
 \textbf{Hongru Liang\textsuperscript{1, 2}\thanks{Corresponding author.}},
 \textbf{Peixin Qin\textsuperscript{1, 2}},
 \textbf{Yao Zhang\textsuperscript{3}},
 \textbf{Wenqiang Lei\textsuperscript{1, 2}}
\\
 \textsuperscript{1}Sichuan University, China \\
 \textsuperscript{2}Engineering Research Center of Machine Learning and Industry Intelligence, \\ Ministry of Education, China \\
  \textsuperscript{3}School of Statistics and Data Science, AAIS, Nankai University, Tianjin, China\\
\href{mailto:xiaoyongjie9@stu.scu.edu.cn}{xiaoyongjie9@stu.scu.edu.cn} \quad \href{mailto:lianghongru@scu.edu.cn}{lianghongru@scu.edu.cn} \quad
\href{mailto:qinpeixin.scu@gmail.com} {qinpeixin.scu@gmail.com} \\
\href{mailto:yaozhang@nankai.edu.cn}{yaozhang@nankai.edu.cn} \quad
\href{mailto:wenqianglei@gmail.com}{wenqianglei@scu.edu.cn}
\\
}

\begin{document}
\maketitle
\begin{abstract}
% Despite the great potential in handling documents and questions, it is still disturbing to fully count on large language models~(LLMs) in real-world searching scenarios. 
Despite the great potential of large language models~(LLMs) in machine comprehension, it is still disturbing to fully count on them in real-world scenarios. 
% This can be partly attributed to the lack of efforts to explain whether the comprehension process of LLMs is aligned with that of humans.
% This is partly due to the limited efforts to explain if LLMs comprehend documents in the same ways humans do.
{This is probably because there is no rational explanation for whether the comprehension process of LLMs is aligned with that of experts.}
In this paper, we propose \name to carefully examine how LLMs perform during the comprehension process from a cognitive view. 
% Specifically, it is equipped with a systematical definition of five required skills during the comprehension process, a strict framework to construct testing data for these skills, and a detailed analysis of off-the-shelf LLMs using the testing data.
Specifically, it is equipped with a systematical definition of five requisite skills during the comprehension process, a strict framework to construct testing data for these skills, and a detailed analysis of advanced open-sourced and closed-sourced LLMs using the testing data.
With \name, we find that it is still challenging for LLMs to perform an expert-level comprehension process. 
Even so, we notice that LLMs share some similarities with experts, e.g., performing better at comprehending local information than global information. 
Further analysis reveals that LLMs can be somewhat unreliable --- they might reach correct answers through flawed comprehension processes. Based on \name, we suggest that one direction for improving LLMs is to focus more on the comprehension process, ensuring all comprehension skills are thoroughly developed during training\footnote{https://github.com/SCUNLP/SCOP}.
\end{abstract}
\section{Introduction}
\begin{figure*}[t]
  \centering
  \includegraphics[width=\textwidth]{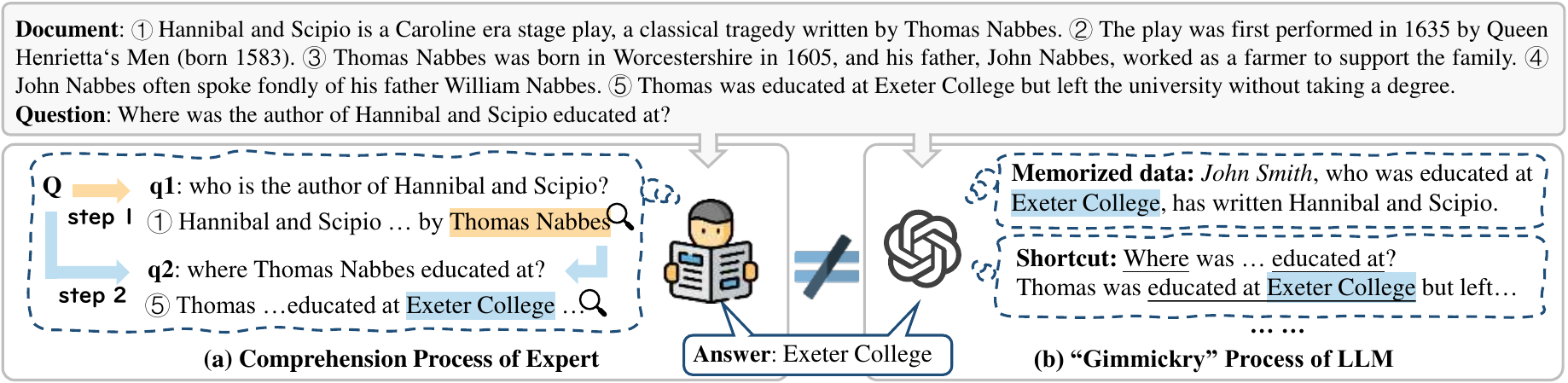}
  \caption{The comprehension processes of an expert and an LLM. With the same document, question, and answer, the expert gets the answer by first identifying who wrote ``Hannibal and Scipio'' from \normalsize{\textcircled{\small{1}}} and then inferring ``Exeter College'' 
 from \normalsize{\textcircled{\small{5}}}, while LLM might use memorized data, shortcuts, etc.}
  \label{intro}
  % \vspace{-5mm}
\end{figure*}

% Large language models (LLMs) have received much attention in the information searching domain because of their ability to cope with massive documents and diverse user questions. Given a document and a question, LLMs are expected to offer the same answer as human experts. 
% Recently, Large language models (LLMs) have received much attention due to their ability of reading comprehension. 
{Large language models (LLMs) have received much attention in acquiring meanings from documents.}
Given a document and a question, LLMs are expected to offer the same answer as human experts.
% Despite great potential, 
However, it is not assured to depend entirely on LLMs in real-world applications~\cite{10.1145/3649506}. One of the possible reasons is that we still don't know whether the comprehension process of LLMs is aligned with experts. 
% As illustrated in Figure~\ref{intro}(a), an expert would answer the question following the process that first identifies who wrote "Hannibal and Scipio" and then locates where that person was educated.
As shown in Figure~\ref{intro}(a), an expert answers the question following the process from step 1 to 2
% \normalsize{\textcircled{\small{1}}} to \normalsize{\textcircled{\small{5}}} 
based on the document.
% that first locates relevant information and then infers.
% \xiaotodo{locates relevant information and then infers}.
% However, LLMs might arrive at the correct answer without such a process while using memorized data or shortcuts shown in Figure~\ref{intro}(b). 
LLMs might arrive at the correct answer following other processes, e.g., using memorized data or shortcuts~( Figure~\ref{intro}(b)).
{This difference
% between experts and LLMs 
doesn't matter in a non-existent scenario where LLMs can give correct answers for every question. However, it matters a lot in real-world safety-critical scenarios~(law, education,  healthcare) especially when normal readers cannot judge the correctness of answers~\cite{pan2023risk, amann2020explainability}. The only solution is to push LLMs to perform the same comprehension process as experts. This encourages us to make a primary attempt and carefully examine the comprehension process of LLMs.}
% Once a LLM generates misinformation using 
% However, it may cause a huge loss in safety-critical domains~(law, education,  healthcare) once the LLM generates misinformation.}
% \xiaotodo{for example, xxx}. 
% \xiaotodo{In fields with low error tolerance, like medicine, this could lead users to place misplaced trust in LLMs, potentially resulting in serious consequences}~\cite{amann2020explainability}.
% This accuracy, sometimes coming from unknown processes shown in Figure~\ref{intro}(b), 
% This high accuracy of responses, despite the unknown process, might cause users to trust LLMs blindly. If LLMs miscomprehend and provide incorrect answers, users may not be able to detect the errors, leading to serious consequences in safety-critical domains~\cite{amann2020explainability}.
% Thus, it necessitates a careful examination of the comprehension process of LLMs.

\par
{While numerous efforts have been made on comprehension evaluation of LLMs, they are busy finding more proper ways to compare the answers generated after the comprehension process with human references~\cite{rajpurkar-etal-2018-know, lai-etal-2017-race, dua-etal-2019-drop}. 
% This may not result in a reliable judgment because higher matching scores do not mean better comprehension of the document~\cite{dunietz-etal-2020-test}. 
This may not provide a reliable judgment, as higher matching scores do not mean better comprehension of the document~\cite{dunietz-etal-2020-test}.
% For example, LLMs can make the same response by relying on shortcuts or retrieving the answer from the memorized training data even without the process of comprehending the document shown in Figure~\ref{intro}(b).
% \xiaotodo{For example, LLMs can make the same response by retrieving the answer from the memorized training data or using shortcuts as shown in Figure~\ref{intro}(b).}
% For example, LLMs can make the same response by using ``gimmickry'' processes shown in Figure~\ref{intro}(b).
A few works~\cite{sugawara-etal-2017-evaluation,wang-etal-2022-feeding,dunietz-etal-2020-test,sugawara-etal-2021-benchmarking} working in the opposite direction --- they prefer to investigate whether LLMs act as accomplished linguists~(such as coreference resolution and named entity recognition) before the comprehension process. 
% They are concerned about the abilities of LLMs in tasks grounding the document, such as coreference resolution, name entity recognition, etc. 
% They focus on the abilities of LLMs in tasks grounding the document, \xiaotodo{such as coreference resolution and named entity recognition}. 
% However, a model, proficient in linguistic abilities like coreference resolution, \xiaotodo{may not be good at completing the comprehension process}~\cite{farr1986reading}. 
% However, a model, proficient in named entity recognition, may not be good at integrating and applying them effectively during the comprehension process~\cite{farr1986reading}.
{However, an LLM, 
% proficient in
good at named entity recognition, may not be good at integrating the whole meaning of a document~\cite{farr1986reading}.}
% \xiaotodo{There is still no method to systematically explain the gap between LLMs and experts during the comprehension process.}
% \xiaotodo{In conclusion, while numerous tasks have been developed to evaluate the comprehension of LLMs, the comprehension process of LLMs remains poorly defined. Consequently, there is no method to systematically explain the gap between LLMs and experts in this regard.}
{Despite various task forms, it remains unclear how LLMs should comprehend the document to close the gap against experts.}
\par
To this end, we propose \name to carefully study whether LLMs have competitive \underline{S}kills with experts during the \underline{CO}mprehension \underline{P}rocess. 
We summarize three fundamental challenges behind \name: 1) a systematical definition of the requisite skills during the comprehension process, 2) a strict framework to construct general testing data for these skills, and 3) a detailed evaluation of LLMs based on the framework. 
% \quad
Towards the definition, we decompose the comprehension process into three levels~(locating, inferring, interpreting) from local to global based on cognition theories~\cite{krathwohl2002revision,afflerbach2015conceptualizing}. The interpreting level is further decomposed into three skills~(connecting, organizing, selecting).
% ~\cite{spivey1990transforming}. 
Finally, the comprehension process comprises five skills~(cf., Table~\ref{tab:tasks}).
% \quad Toward the framework, we introduce a series of strict rules to make sure the testing data focuses on the comprehension process without distraction factors.
% For example, we have filtered all questions~(e.g., the question in Figure~\ref{intro}) that don't need LLMs to comprehend the document, namely, those that can be correctly answered by LLMs without referring to any documents. 
% \quad 
% \xiao{Toward the framework, we introduce a series of strict rules to make sure the testing data focuses on the comprehension process without distraction factors. 
% These rules are established to deal with general data, regardless of its specific format.
% We argue that the comprehension evaluation should not being influenced by document types or answer styles. 
% For the generality of testing data, we involves both narrative~(mainly describing events, like news) and expository documents~(mainly describing facts, like encyclopedias) with diverse answer styles for each task.} 
{Towards the framework, we argue the testing data should be independent of formats~(e.g., document types, answer styles) and merely focus on the comprehension process. In the absence of suitable data, we introduce a series of strict rules to modify existing datasets and crawl new datasets based on the definitions of skills. The final testing data includes 4,682 samples from 12 datasets.}
% In the absence of a suitable dataset, we construct the testing data by modifying the existing datasets and crawling new datasets according to the definitions. The final testing data consists of 4,682 samples from 12 datasets. 
% \quad Toward the evaluation, we compare the performances of two close-sourced~\cite{yang2024qwen2,meta2024llama} and two open-sourced~\cite{anthropic2024claude,openai2024} LLMs on different levels, skills, document types, and answer styles. 
% \quad 
Towards the evaluation, we compare the performances of two close-sourced and two open-sourced LLMs on different levels, skills, document types, and answer styles. 
\par

With \name, we find that no model has yet achieved the expert-level comprehension process. 
% Besides, aligned with humans ~\cite{kintsch1978toward}, LLMs are better at local comprehension than global comprehension -- all LLMs perform better at the locating level than at the inferring and interpreting levels. 
Besides, aligned with {human results}~\cite{kintsch1978toward}, LLMs are better at local comprehension than global comprehension --- better at the locating level than at the inferring and interpreting levels.
Surprisingly, our findings with \name differ from existing LLM evaluations~\cite{wang2024mmlu,zheng2023judging}, where larger closed-sourced models usually outperform open-sourced ones, except for the locating skill. 
This suggests that, similar to humans, LLMs do not gain better comprehension just by memorizing more information. 
% We also observe variability in the relative performance of different models. For example, Llama~\cite{meta2024llama} ranks first on the inferring skill yet doesn't work well on the selecting skill. 
% \xiaotodo{Therefore, one potential improvement for LLMs is to thoroughly learn all comprehension skills during training. Further analysis shows that LLMs may guess correctly from an incorrect comprehension process.
% % , e.g., answering the question in Figure~\ref{intro} only based on sentence~\normalsize{\textcircled{\small{2}}}. 
% Moreover, if the comprehension process goes correctly, all LLMs experience a big increase in the inferring skill. This underscores the importance of having a correct comprehension process in LLMs. }
% {Evaluating comprehension process provides a {promising and believable} measure of LLMs' comprehension from a cognitive view.}
{Further analysis shows that if the comprehension process goes correctly,
% ~(e.g. the process in Figure~\ref{intro}(a)), 
all LLMs experience a big increase in the inferring skill.} One potential improvement for LLMs is to thoroughly learn all comprehension skills during training.} {We believe evaluating comprehension process provides insightful observations about machine comprehension and gives a new perspective to motivate the future study of LLMs.}
We hope \name can shine a light on how to comprehend documents like experts, accelerating the reliable deployment of LLMs in safety-critical applications. Our contributions are as follows.\par

\begin{itemize}
\item We emphasize the gap between the comprehension process of LLMs and experts, which slows down the real-world application of LLMs.
\item We propose \name to explore the comprehension process of LLMs from a cognitive view. It includes a systematic definition of five requisite comprehension skills, a strict data construction framework, and a detailed analysis of LLMs.
% is equipped with 

\item With \name, we provide insights into how LLMs perform during the comprehension process, 
% \xiaotodo{facilitating future research in the areas of document comprehension and model design.}
{facilitating future research in the improvement and deployment of reliable models.}
\end{itemize}

\begin{table*}[t]

\centering
\scalebox{0.7}{
\begin{tabular}{lm{9.5cm}m{5cm}m{5cm}}
\toprule
\textbf{Skill} & \textbf{Task Description} & \textbf{Input} & \textbf{Output}  
\\ \hline
 Locating & identify a sentence that supports answering the question & a document and a question & {a supporting sentence and answer}
\\ \hline
Inferring & identify sentences that support answering the question &  a document and a question & supporting sentences and answer
\\ \hline
 Connecting & choose a sentence to connect the previous and following context  & a document with blanks and sentence candidates & sentences selected to fill blanks
\\ \hline
    Organizing & organize the document based on the subheadings & a document with position sequence numbers and subheadings & the positions for subheadings in the document
\\ \hline
    Selecting  & select the key sentences of the document & a document and the desired number of key sentences & key sentences
\\ \bottomrule
\end{tabular}
}
\caption{{Definitions of the tasks designed for each skill. Data examples are presents in Appendix~\ref{data_examples}.}}
\label{tab:tasks}
\end{table*}
\section{Task Definition}
% According to ~\citet{afflerbach2015conceptualizing}, comprehension can be divided into basic skills (i.e. access text-explicit meanings) and higher-order thinking (i.e. grasp the text using text information and internal knowledge). In this paper, we only focus on these basic skills about comprehension process, as the internal knowledge of an LLM is related to its training data rather than its comprehension ability.
{The ultimate goal of reading comprehension is to reconstruct information in a document to a meaningful representation in mind and apply it in new situations~\cite{Kintsch1998ComprehensionAP}. 
% \xiaotodo{According to ~\citet{afflerbach2015conceptualizing}, comprehension can be divided into basic skills (i.e. access local and global text-explicit meanings) and higher-order thinking (i.e. grasp the text using text information and internal knowledge). 
% Since the internal knowledge of an LLM is related to its training data rather than its comprehension ability, we only focus on these basic skills about comprehension process. }
{According to ~\citet{afflerbach2015conceptualizing}, this can be achieved by two processes: the comprehension process, where the reader must obtain local and global meanings of the document; and the thinking process, where the reader must combine these meanings with his background knowledge. In this paper, we focus on the comprehension process, as the gap in background knowledge between LLMs and experts can be narrowed down by feeding more data.}
From a cognitive view~\cite{krathwohl2002revision,afflerbach2015conceptualizing}, we decompose the comprehension process into three levels from local to global and propose five skills. Inspired by educational practices~\cite{van2012connecting, carrell1998can}, we design five tasks specifically tailored to evaluate each skill, as explained in Table~\ref{tab:tasks}. The detailed definitions are as follows.\par
\paragraph{Locating} The locating level involves questions that should be answered by a piece of information in the document. We treat the sentence containing such information as a supporting sentence. The locating skill focuses on phrase- or sentence-level information and is the basic skill of the comprehension process. 
% \xiao{To evaluate it, we define a task --- before answering a question, LLMs must identify the sentence containing the information that supports their answer. The identified sentence refers to the supporting sentence.}}
To evaluate it, we define a task where the LLM needs to give the supporting sentence and the answer to a question based on the document.
\paragraph{Inferring} The inferring level involves questions that should be answered by multiple pieces of information in the document.
% e.g., the question in Figure~\ref{intro}. 
{The inferring skill is more challenging than the locating skill, focusing on sentence- or discourse-level information. To evaluate it, 
% similar to the locating skill, 
we define a task where LLMs need to identify multiple supporting sentences before answering a question. For example, LLMs should identify \normalsize{\textcircled{\small{1}}} and \normalsize{\textcircled{\small{5}}}, and then answer with ``Exeter College'' to the question in Figure~\ref{intro}.}

{\paragraph{Interpreting} The interpreting level involves questions that should be answered by the whole content of a document. This is very similar to making a summary of the document. Instead of directly using the automated summarization task, we care more about the comprehension process before getting the final summary. According to \citet{spivey1990transforming}, this process can be decomposed into the connecting, organizing, and selecting skills:}
\begin{itemize}
    % \item With the connecting skill, one can connect discrete messages in a document sentence by sentence. It is a bit like the next sentence prediction task~\cite{devlin2019bert}, where LLMs are required to predict the next sentence given the previous context. We prefer the sentence cloze task, a more suitable task where LLMs are required to connect the sentence with both the previous and following context.
    \item With the connecting skill, one can connect discrete messages in a document sentence by sentence. It is a bit like the next sentence prediction task~\cite{devlin2019bert}, where LLMs should decide whether two sentences appear consecutively in a document. 
    % where LLMs are required to find the next sentence based on the previous context. 
    We prefer the sentence cloze task, a more suitable task where LLMs must determine the correct sentence to connect both the previous and the following context.
    % \item With the organizing skill, one can separate the document into several meaningful chunks in logical ways. Specifically, we design a text segmentation task, where LLMs are required to organize the documents based on given subheadings.
    \item With the organizing skill, one can separate the document into several meaningful chunks in logical ways. We design a text segmentation task, where LLMs are required to organize documents based on given subheadings.
    % \item With the selecting skill, one can construct the meaning of the document with several key sentences. We evaluate this skill by asking the LLMs to extract key sentences from the document.
    \item With the selecting skill, one can capture the meaning of a document with several key sentences. We evaluate this skill by tasking LLMs to extract key sentences of the document.
\end{itemize}

\section{Data Construction Framework} 
Despite numerous efforts to evaluate comprehension, there remains a lack of suitable testing data to evaluate the comprehension process. With the above definitions, we introduce a data construction framework with a series of strict rules to ensure that the testing data reliably evaluates each skill without distractions. Besides, for general testing data, {we include both narrative and expository document types, along with various answer styles. }
% The final dataset, as detailed in Table~\ref{tab:data stastics}, comprises 4,682 samples sourced from both existing and newly created datasets.

\subsection{Locating}
\label{sec:loc}
We define the locating question as one that focuses on factual details or events explicitly presented in the document. We consider a sentence as the basic unit that encapsulates a fact or an event. Therefore, locating questions can be answered by referencing a single supporting sentence within the document. \par
We utilize three existing datasets as our source data, each containing questions focused on facts or events: SQuAD v2.0~\cite{rajpurkar-etal-2018-know}, an expository-type dataset with spans extracted from documents as answers; NewsQA~\cite{trischler-etal-2017-newsqa}, a narrative-type dataset also with span-style answers; and MCTest~\cite{richardson-etal-2013-mctest}, a narrative-type dataset with multiple-choice answers. 
% \xiao{Note that, despite the answer is provided, the task is also required finding a supporting sentence.} 
Fortunately, SQuAD v2.0 and NewsQA provide annotated spans, making it natural to consider sentences containing these spans as supporting sentences. 
{A direct solution is to treat the questions with only one annotated supporting sentence as locating questions.}
% \xiao{However, a manual check shows that the annotation may contain the span but not support to answer the question. For example, for the question ``Who wrote Hannibal and Scipio?'' with the answer ``Thomas Nabbes'', \normalsize{\textcircled{\small{3}}} in Figure~\ref{intro} doesn't support the answer.}
{However, we find this may bring much noise. Some annotated spans do contain the same tokens of answers but are not related to supporting sentences. In Figure~\ref{intro}, sentence \normalsize{\textcircled{\small{3}}} contains ``Thomas Nabbes'' but does not support to answer $q_1$.}
% , cf. sentence~~\normalsize{\textcircled{\small{{1}}}} marked with an $\times$ symbol in the upper part of Figure~\ref{asnwer_not_support}. 
Even worse, MCTest only annotates the correct answers, which may not be exactly presented in the document. Thus, the challenge lies in annotating the supporting sentences for questions.
\par

We employ a simple method based on semantic similarity to select the candidate supporting sentences. 
% Specifically, if a sentence offers enough information to answer the question, we deem it a candidate supporting sentence. 
To find such candidates, the question and its corresponding answer are first transformed into a declarative sentence by prompting Llama3-8B-Instruct~\cite{meta2024llama3}. For example, the question ``Who's the author of Hannibal and Scipio?'' and its answer ``Thomas Nabbes'' are transformed into ``Thomas Nabbes is the author of Hannibal and Scipio.'' This declarative sentence~($d$) is then used to retrieve candidate supporting sentences by computing a z-score score with the $i^{th}(0<i\leq |\mathbb{D}|)$ sentence in the document~($\mathbb{D}$):
\begin{equation}
\resizebox{0.85\linewidth}{!}{%
\( z_i(d) = \frac{S(d, s_i)-\mu(S(d, s_1),\dots,S(d, s_{|\mathbb{D}|}))}{\sigma(S(d, s_1),\dots,S(d, s_{|\mathbb{D}|}))} \)%
}
\end{equation}
where $S(\ast, \star)$ calculates the cosine similarity between $\ast$ and $\star$, $\mu(\bullet)$ and $\sigma(\bullet)$ calculates the mean and deviation of $\bullet$, respectively, more details are shown in Appendix~\ref{semantic}. The candidate supporting sentences are ones with z-scores above pre-defined thresholds\footnote{The threshold varies by datasets, we set it to 1 for SQuAD v2.0 and 3 for NewsQA.}.
%[/Move to appendix]
{We notice this retrieval may be sensitive to answers, particularly answers appearing only once in the document. Thus, we utilize both declarative sentences and questions to retrieve sentences. The final candidate set is the intersection of these retrieval results.}
%[/Move to appendix]
\par
{For datasets with annotated spans, if the annotated supporting sentence is in the candidate set, we treat the question as a locating question.}
% We refer to this method as the first solution}
% , illustrated in the first solution in Figure~\ref{asnwer_not_support}. 
% This approach enables us to filter about 40\% of questions with one supporting sentence in SQuAD 2.0 and over 50\% in NewsQA.
% \xiao{For datasets without annotated spans, we use the answer to retrieve a pseudo-supporting sentence. We treat this sentence as the annotated supporting sentence mentioned in the first solution and follow the same process to identify the locating question.}
{For datasets without annotated spans, we use the answer to retrieve a pseudo-supporting sentence. If the retrieved sentence is in the candidate set, we treat the question as a locating question.}
% , as shown in the second solution in Figure~\ref{asnwer_not_support}. 
%{We observe that $S(\ast, \star)$ doesn't work well when the answer is too short to be 
% well-encoded in embedding spaces.embedded.As such, the pseudo-supporting sentence is retrieved by BM25~\cite{robertson1994some}.}
%[Move to appendix]
% \xiao{To verify the pseudo solution, we used annotated datasets SQuAD 2.0 and NewsQA. The locating questions identified from the first solution are treated as the gold results, and the results of the pseudo solution are compared to them. In a classification setting, the pseudo solution achieves a 0.94 F1 score on SQuAD 2.0 and a 0.91 F1 score on NewsQA, demonstrating its consistency with the first solution.}
{To verify the pseudo solution, we also apply it on SQuAD v2.0 and NewsQA. We use the questions identified from annotated sentences as gold results and the questions identified from the pseudo solution as predicted results. The F1 scores of the pseudo solution are 0.94 on SQuAD v2.0 and 0.91 on NewsQA.}

%[/Move to appendix]

% \begin{figure}[t]
%   \centering
%   \includegraphics[width=\linewidth]{Figure/locating_solution.pdf}
%   \caption{The two solutions for identifying locating-type questions. The first solution is used to handle datasets with answer annotations, while the second handles datasets without them.}
%   \label{asnwer_not_support}
% \end{figure}

% \begin{figure}[t]
%   \centering
%   \includegraphics[width=\linewidth]{Figure/bridge_infer.pdf}
%   \caption{Illustrations of a non-inferring question~(upper) and inferring question~(lower)}
%   \label{bridge}
% \end{figure}

% \begin{figure}[t]
%   \centering
%   \includegraphics[width=0.88\linewidth]{Figure/com.png}
%   \caption{Illustrations of a non-inferring question~(upper) and inferring question~(lower), presenting the comparison structure.}
%   \label{comparison}
% \end{figure}

\subsection{Inferring}
\label{sec:inf}
We define the inferring question as one that should be answered by integrating multiple supporting sentences. We consider three datasets as our source data: HotpotQA~\cite{yang2024qwen2}, an expository-type dataset with span-style answers; MusiQue~\cite{trivedi-etal-2022-musique}, also an expository-type dataset with span-style answers; and RACE~\cite{lai-etal-2017-race}, a narrative-type dataset with multi-choice answers.
\begin{figure}[t]
  \centering
  \includegraphics[width=\linewidth]{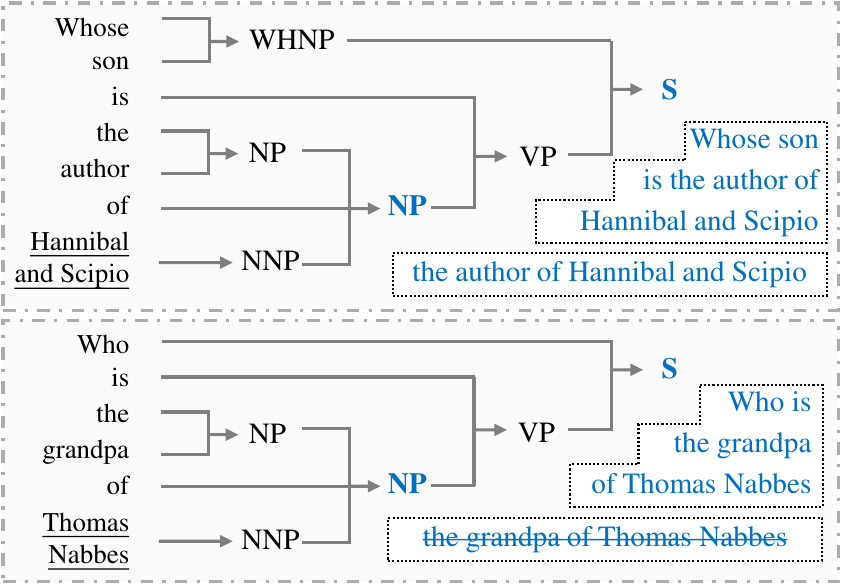}
  \caption{The syntax trees of an inferring question~(upper) and non-inferring question~(lower).}
  \label{fig:bridge}
\end{figure}
Although each question is annotated with supporting sentences in HotpotQA and MusiQue, not all questions are inferring questions. 
% Taking Figure~\ref{bridge} as an example, both questions are annotated with two supporting sentences. However, the upper question requires commonsense knowledge beyond the document to answer.  
% We suppose both questions can be decomposed into subquestions, each corresponding to a supporting sentence. Only the lower question -- an inferring question -- can be decomposed into more than one subquestions using semantics alone, without additional knowledge. 
% \xiao{For example, consider the question ``Who is the grandpa of Thomas Nabbes?''. While this question requires two supporting sentences (\normalsize{\textcircled{\small{3}}} and \normalsize{\textcircled{\small{4}}} in Figure~\ref{intro}), it requires commonsense knowledge to answer.} 
{For example, to answer ``Who is the grandpa of Thomas Nabbes?'', it requires not only \normalsize{\textcircled{\small{3}}} and \normalsize{\textcircled{\small{4}}} from the document in Figure~\ref{intro} but extra commonsense knowledge.}
To remove such questions, we suppose that an inferring question can be decomposed into more than one subquestion
% based on semantics alone, without additional knowledge.
{without additional knowledge beyond the document.}\par
% We design a semantic-based decomposition method to identify inferring questions by judging the number of subquestions. 
{Thus, for datasets with annotated supporting sentences, we design a semantic-based method to decompose the question into subquestions and identify an inferring question by the number of its subquestions.}
Specifically, we use the Berkeley Neural Parser\footnote{https://spacy.io/universe/project/self-attentive-parser/} to divide the questions into syntax trees. These trees represent how a linear string of words in a question connects to its meaning~\cite{caplan1988disorders} and help identify whether the question consists of multiple subquestions. 
We obtain the subquestion candidates by pruning the edges of the ``NP'' node, at least one of whose child nodes contains a named entity.
The question is also included as a root subquestion to prevent any loss of information. In this way, we obtain two candidates for the question in Figure~\ref{fig:bridge}~(upper). We note that different candidates may target the same fact and represent the repetitive subquestion, e.g., the two candidates in Figure~\ref{fig:bridge}~(lower). To filter out such candidates, we compute the cosine similarity distribution of each candidate over the document sentences. We compare the correlation between the distributions of any candidate pairs. If the correlation coefficient is greater than 0.8\footnote{We choose 0.8 as it is commonly used to represent a strong correlation in statistical practice.}, the candidate with a deeper position in the syntax tree is removed. This method helps us identify inferring questions from bridge questions, where each subquestion contains only a single entity. However, it cannot handle comparison questions, which have subquestion candidates with multiple entities. We identify these questions using their syntactic structure where conjunctions~({e.g., ``and''}) connect entities. Some questions~(e.g., ``Who is older, Queen's Men or Thomas Nabbes) with comparative words like ``older'' require mathematical knowledge beyond the document. We further utilize part-of-speech tagging ``JJR'' and ``RB'' to filter out such questions.

For datasets without annotated supporting sentences~(RACE), we prompt GPT-4o-mini~\cite{openai2024mini} to retrieve supporting sentences. Since LLMs might rely on their inherent knowledge, we also perform retrieval based on semantic similarity\footnote{A case showing how LLMs rely on their inherent knowledge to retrieve is provided in Appendix~\ref{app:case}.}. We decompose the declarative sentence into sub-sentences, following the same constraints of subquestions. These sub-sentences are then used to retrieve supporting sentences via the solutions proposed in Section~\ref{sec:loc}. The final supporting sentences must appear in both the retrieved sets. At last, we identify inferring questions by remaining questions with more than one supporting sentence. 
{To showcase the effectiveness of our method for extracting supporting sentences from narrative-style data, we test on two small-scale datasets with annotated supporting sentences: MultiRC~\cite{khashabi-etal-2018-looking} and OnestopQA~\cite{berzak-etal-2020-starc}. The F1 scores for supporting sentence identification are 0.85 and 0.96, respectively, indicating that our method is highly reliable.}

\subsection{Connecting}
\label{sec:con}
We introduce the sentence cloze task to evaluate the connection skill, where LLMs are required to fill in blanks in the document with a set of candidate sentences. 
% \xiaotodo{We find SCDE ~\cite{kong-etal-2020-scde}, a sentence cloze dataset containing narrative-type documents from public school examinations, is suitable for this task.} 
{We use SCDE~\cite{kong-etal-2020-scde} as our narrative-type dataset and create a new sentence cloze dataset for expository-type data.}
% Additionally, we create a new expository-type sentence cloze dataset. 
Specifically, we extract all ``Introduction'' sections from the {ACL OCL Corpus~\cite{rohatgi-etal-2023-acl}} as documents. Inspired {by~\citet{cui-etal-2020-sentence}}, we randomly select five sentences from each document as clozes.
% , ensuring they are not the first or last sentences, nor consecutive. 
{These clozes are not consecutive and are not the first or last sentences of a document.}
We also generate two distractions for each document by randomly sampling candidate sentences from other sections of the paper. After computing the cosine similarity scores with all clozes, we take the top-2 candidates with $<0.6$ scores as distractions.

\begin{table}[t]
\scalebox{0.72}{
\begin{tabular}{lcccc}
\toprule
\textbf{Skill} & \textbf{Source} & \textbf{Size} & \textbf{\makecell[c]{Answer \\ style}} & \textbf{\makecell[c]{Document\\ type}} \\ \hline
\multirow{3}{*}{Locating}   
& SQuAD v2.0  & 479 & Spans & Expository 
\\ \cline{2-5} 
& NewsQA & 984 & Spans & Narrative
\\ \cline{2-5} 
& MCTest & 72 & Multi-choice & Narrative              
\\ \hline
\multirow{3}{*}{Inferring}  
& HotpotQA & 604 & Spans & Expository 
\\ \cline{2-5} 
& MusiQue & 510 & Spans & Expository
\\ \cline{2-5} 
& RACE & 547 & Multi-choice & Narrative
\\ \hline
\multirow{2}{*}{Connecting} 
& SCDE & 625 & - & Narrative
\\ \cline{2-5} 
& ACL OCL & 169 & - & Expository
\\ \hline
\multirow{2}{*}{Organizing} 
& ClimateCentral & 108 & - & Narrative 
\\ \cline{2-5} 
& WikiHow & 146 & - & Expository
\\ \hline
\multirow{2}{*}{Selecting}  
& SourceSum & 143 & - & Narrative
\\ \cline{2-5} 
& ACL OCL & 295 & -  & Expository             
\\ \bottomrule
\end{tabular}}
\caption{The statistics of our testing data.}
\label{tab:data stastics}
% \vspace{-5mm}
\end{table}

\subsection{Organizing}
% \xiao{To evaluate the organizing skill, we introduce a new text segmentation task. In this task, LLMs should insert the subheadings into the document to organize it into several subsections. Existing setting requires LLMs to directly predicts segmentation positions in a given document. This approach may lead to evaluation errors since an document can be segmented in various valid ways. By adding subheadings as semantic constraints, we ensure consistent segmentation results. Besides, our task format is similar to next-sentence prediction, making it more suitable for LLMs to handle. We review existing datasets for text segmentation~\cite{koshorek-etal-2018-text, arnold2019sector} including subheadings. These subheadings mainly offer structural information like ``Preface'', rather than semantic information. Thus, we collect data from WikiHow~\cite{Koupaee2018WikiHowAL} as an expository dataset. We also crawl news documents from Climate Central\footnote{https://www.climatecentral.org/what-we-do/legal} as a narrative dataset. We ensure each subheading has at least 4 words to provide enough semantic information.}
The organizing skill is evaluated via a text segmentation task. Instead of just predicting segmentation positions, we require LLMs to organize a document based on given meaningful subheadings. With this constraint, LLMs are more likely to provide consistent segmentation results. 
% We also want the setting to be similar to next sentence prediction, which is one of the most familiar tasks for LLMs. 
% We have noticed two datasets~\cite{koshorek-etal-2018-text, arnold2019sector} meet the required format. However, their subheadings mostly refer to structures~(e.g., ``Preface'') without enough semantic meanings. 
% {We have noticed two text segmentation datasets~\cite{koshorek-etal-2018-text, arnold2019sector} meet the required format. However, their subheadings mostly refer to structures~(e.g., ``Preface'') without enough semantic meanings.}
While existing text segmentation datasets~\cite{koshorek-etal-2018-text, arnold2019sector} meet the required format, their subheadings mostly refer to structures~(e.g., ``Preface'') without enough semantic meanings.
{For narrative-type data, we crawl news documents with subheadings from Climate Central\footnote{https://www.climatecentral.org/what-we-do/legal}. For expository-type data, we modify the WikiHow~\cite{Koupaee2018WikiHowAL} dataset by forming related paragraphs as documents and using summaries as subheadings.}
% Thus,  we collect documents with meaningful subheadings from Climate Central\footnote{https://www.climatecentral.org/what-we-do/legal} and WikiHow~\cite{Koupaee2018WikiHowAL} to construct our narrative and expository dataset, respectively.
% \xiaotodo{Thus, we collect expository-type data from WikiHow~\cite{Koupaee2018WikiHowAL} which provides subheadings as the summary for a document.} For narrative-type data, We crawl news documents from Climate Central\footnote{https://www.climatecentral.org/what-we-do/legal}.
% \xiaotodo{we collect documents from WikiHow~\cite{Koupaee2018WikiHowAL} as an expository dataset} and documents from Climate Central\footnote{https://www.climatecentral.org/what-we-do/legal} as a narrative dataset. 
Each subheading must contain at least four words to serve as a segmentation constraint.

\begin{table*}[t] 
\resizebox{\linewidth}{!}{
\begin{tabular}{l|ccc|ccc|cc|cc|cc}
\toprule
\multirow{2}{*}{\textbf{Model}} & \multicolumn{3}{c|}{\textbf{Locating}}                 & \multicolumn{3}{c|}{\textbf{Inferring}}              & \multicolumn{2}{c|}{\textbf{Connecting}} & \multicolumn{2}{c|}{\textbf{Orgnazing}}    & \multicolumn{2}{c}{\textbf{Selecting}} \\ \cline{2-13} 
                                    & \textbf{MCTest} & \textbf{SQuAD} & \textbf{NewsQA} & \textbf{HotpotQA} & \textbf{MusiQue} & \textbf{RACE} & \textbf{ \ SCDE \ }     & \textbf{ACL}     & \textbf{Climate} & \textbf{WikiHow} & \textbf{SourceSum}  & \textbf{ACL} \\ \hline
Qwen2-72B-Instruct              & 97.22           & 98.33              & 84.15           & 46.19             & 11.37            & 37.29         & 72.48             & 33.14                & 12.96                   & 22.60            & 16.08               & 9.83             \\
Llama3.1-70B-Instruct           & 94.44           & 99.16              & 85.67           & 62.58             & 30.98            & 46.07         & 38.88             & 11.83                & 25.93                   & 34.93            & 16.78               & 9.83             \\
% GPT-4o-mini                     & 91.67           & 97.49              & 85.87           & 34.44             & 11.37            & 37.84         & 68.25             & 10.06                & 12.04                   & 30.82            & 10.49               & 12.54            \\
Claude-3.5-sonnet               & 97.22           & 97.08              & 84.04           & 56.13             & 24.71            & 31.63         & 69.12             & 60.95                & 43.52                   & 55.48            & 18.18               & 10.17            \\
GPT-4o                          & 97.22           & 98.96              & 89.13           & 45.36             & 18.82            & 37.48         & 58.72             & 24.85                & 27.78                   & 35.62            & 22.38               & 12.54            \\ \bottomrule
\end{tabular}}
% \vspace{-5mm}
\caption{Performance of model's comprehension process across five skills.} 
\label{tab:overallperformance}
\end{table*}
\subsection{Selecting}
% \xiao{We introduce a new task called key sentence selection to evaluate the selecting skill. In this task, LLMs are required to extract key sentences that capture the main idea of a document. 
% We adapt the SourceSum~\cite{suhara-alikaniotis-2024-source} dataset as narrative-type data source. It is originally designed for identifying source sentences of summaries from news documents. These human-annotated source sentences serve as our gold key sentences. For expository-type data, we use ``Introduction'' sections from ACL papers in the ACL OCL corpus as our document source. Since the ``Abstract'' typically summarizes the ``Introduction'', it serves as a reference for identifying key sentences. We then match each abstract sentence with the most similar sentence from the ``Introduction'' and label them as key sentences.}

{To evaluate the selecting skill, we leverage a key sentence selection task, where LLMs must extract several key sentences from the document to condense its main idea. For narrative-type data, we adopt SourceSum~\cite{suhara-alikaniotis-2024-source}, which equips each document with a summary and its source sentences. We treat the source sentences as golden key sentences. For expository-type data, we reuse the ``Introduction'' data collected in Sec.~\ref{sec:con}. We further collect the ``Abstract'' sections for the ACL OCL Corpus~\cite{rohatgi-etal-2023-acl}. Each ``Abstract'' sentence is assigned with the most similar ``Introduction'' sentence. We treat these assigned sentences as golden key sentences.}

\subsection{Quality Control and Analysis}
\label{quality}
The five tasks are designed to ensure LLMs genuinely comprehend the documents when answering, preventing scenarios like ``shortcuts''.
To address cases where LLMs rely on their memory, we further exclude questions that can be directly answered by advanced LLMs~(Llama3.1-70B-instruct~\cite{meta2024llama} and GPT-4o~\cite{openai2024}) without referring to the document\footnote{This step is deemed necessary, as noted in \cite{chang2023speak} and validated through our pilot experiment in Appendix~\ref{Memorized_exp}.}. 
Specifically, for datasets with span-style answers, we convert the question into a yes-no question by merging its answer. Then, the LLMs are required to answer this new question. We only keep the question if both answers are ``no''. For multi-choice datasets, we reuse the strategy proposed in Sec.~\ref{sec:loc} and transform the question with its options into a set of declarative sentences. Then, we require LLMs to select the most sensible one from these declarative sentences. We only keep questions if both selected sentences do not contain the right option. 
% In this way, we filter out 33,827 noise samples from 37,023 samples.} {We are surprised that such a large proportion of test data have been memorized by LLMs.}
Surprisingly, we filter out more than 90\%~(33,827) noise samples from 37,023 samples. This further indicates that it is still unsafe to directly deploy LLMs in real-world scenarios, particularly when domain knowledge has conflicts with pre-trained corpora.
% that a large proportion of questions can be answered by LLMs using memorized data, 

During the data construction phase, the API cost is only about \$5, including about \$0.9 for supporting sentence retrieval and about \$4.1 for filtering.
% \xiao{The total cost is only about \$5, including about \$0.9 for supporting sentence retrieval and about \$4.1 for filtering.}
Additionally, we randomly select 50 samples from each dataset and ask three workers to check the {validity} of each sample. A sample gets a score of 1 if all workers agree it is a valid sample; otherwise, the sample is scored 0. {The average score achieves 0.81 with an inter-annotator agreement score of 0.73}. The statistics of our testing data are shown in Table~\ref{tab:data stastics}. More details are provided in the Appendix~\ref{collection_detail}.
% \xiao{
% After collecting the testing data, we exclude QA data that LLMs can answer without context to avoid distractions. For datasets with span-style answers, we convert questions and answers into yes-no questions. 
% For example, the question in Figure~\ref{intro} becomes, ``Was Thomas Nabbes the author of Hannibal and Scipio?''. 
% We discard questions LLMs can answer ``yes'' to. 
% For multi-choice datasets, we merge each question with its options to create a set of declarative sentences. 
% If LLMs can correctly identify the declarative sentence matching the correct option, we discard that question. We use advanced LLMs for filtering, including the open-source model Llama3.1-70B-instruct~\cite{meta2024llama} and the closed-source model GPT-4o~\cite{openai2024}. Only data validated by both models is retained. Finally, we filter 3196 test samples from six QA datasets, originally totaling 37,023 samples, achieving a data retention rate of 8.6\% at a minimal cost of just \$4.06. Additionally, we randomly select 50 samples from each dataset and have three workers check the validity of the data. The valid rate is 0.81 with an inter-annotator agreement score of 0.73. The testing data statistics are shown in Table~\ref{tab:data stastics}. More details are provided in the Appendix~\ref{collection_detail}.}

\section{Results and Analysis}
\subsection{Settings}
\paragraph{Evaluated Models} 
We involve four LLMs that excel in comprehending documents in \name: two open-sourced models~(Llama3.1-70B-Instruct~\cite{meta2024llama} and Qwen2-72B-Instruct~\cite{yang2024qwen2}) that are deployed directly for inference and two closed-sourced models~(GPT-4o~\cite{openai2024} and Claude-3.5-sonnet~\cite{anthropic2024claude}) that return answers through API calls\footnote{We also evaluate DeepSeek-R1~\cite{guo2025deepseek} and GPT-4o-mini, whose performances are consistent with existing results. For LLMs of varying scales, Llama 3.1 7B is tested but excluded from the main evaluation due to its extremely poor performance. More details are in the Appendix~\ref{app:overll}.}.
% To investigate LLMs' performance in the comprehension process, we select those that excel in comprehending documents. We evaluate LLMs of two categories: (1) Open-source models that are deployed directly for inference, including Llama3.1-70B-Instruct~\cite{meta2024llama} and Qwen2-72B-Instruct~\cite{yang2024qwen2}; (2) Closed-source models that return answers through API calls, including GPT-4o~\cite{openai2024} and Claude-3.5-sonnet~\cite{anthropic2024claude}. 
We set the temperature for all LLMs to 0 
% (i.e., greedy decoding) 
to ensure controllable results.
% and reproducible results. 
The prompts for all tasks are listed in the Appendix~\ref{prompts}.
\paragraph{Metrics} 
{We use document-level accuracy~\cite{yessenalina-etal-2010-multi} as the metric for evaluating all skills.}
% To compare LLMs' performance across different kills, we use document-level accuracy as the metric. 
{
% The accuracy is defined as the number of times LLM's response exactly matches the golden answer for a sample, divided by the dataset's total sample size.
The accuracy is defined as the number of times LLM's predictions exactly match the golden results, divided by the total sample size.
For the locating and inferring skill, we only focus on the predicted supporting sentences as they reflect the comprehension process of LLMs.} The connecting skill requires predicting candidate sentences in the correct sequence. The organizing skill requires predicting the correct positions for subheadings within a document. For the selecting skill, LLMs should predict key sentences.

\subsection{Overall Performance}
We report the performances of LLMs on five skills in Table~\ref{tab:overallperformance}. It can be seen that all LLMs are far from operating expert-level comprehension processes.
%, especially in the selecting skill.
% , as shown in Figure~\ref{radar}. 
Besides, we have the following observations:\par

% \begin{figure}[t]
%   \centering
%   \includegraphics[width=0.7\linewidth]{Figure/radar.pdf}
%   \caption{\xiaotodo{Average scores of LLMs on five skills.}}
%   \label{radar}
%   % \vspace{-5mm}
% \end{figure}

\paragraph{Performance w.r.t., different levels} 
LLMs are better at local comprehension than global comprehension. Their performance decreases significantly as the comprehension process moves from local to global levels. Specifically, the average accuracy drops from 93.55\% at the locating level to 37.38\% at the inferring level, and further to 31.02\% at the interpreting level. This suggests while LLMs handle local comprehension well, they struggle with global comprehension. This limitation may result from the next-token prediction task used during pretraining. This training paradigm excels at capturing local contextual information, but it might hinder the model's ability to grasp the bigger picture.
% All LLMs excel at the locating level, achieving over 95\% accuracy on SQuAD and MCTest. However, their performance drops significantly at the inferring and interpreting levels, with average accuracy falling from 93.55\% at the locating level to 37.38\% at the inferring level, and further down to 31.02\% at the interpreting level. These three levels represent a progression in the comprehension process, moving from local to global comprehension. This progression suggests that LLMs excel in dealing with local comprehension but struggle with global comprehension. This limitation might stem from the next-token prediction task employed during pretraining. While this training paradigm excels at capturing local contextual information, it potentially impedes the model's capacity to grasp the bigger picture.

\paragraph{Performance w.r.t., different skills at the interpreting level} 
% From a broader perspective, 
The three skills at the interpreting level focus on sentences, then discourse, and finally the whole document. The LLMs' performance declines across the three skills, supporting the observation that LLMs are better at local comprehension than global comprehension. For the individual connecting skill, all LLMs perform over twice as poorly on the SCDE dataset compared to the ACL dataset except Claude-3.5. This suggests that most LLMs handle sentence-level logical relationships in narratives better than in expository documents. Besides, LLMs have difficulty with the selecting skill, achieving an average accuracy of only 18.36\%.  This is likely because the task requires focusing on key sentences spread throughout the document, conflicting with the sequential text processing approach of LLMs\footnote{A case study of connecting skill is listed in Appendix~\ref{app:case}.}. 

\paragraph{Performance w.r.t., different LLMs}
Surprisingly, we notice that the performance ranking of different LLMs on \name differs from typical LLMs evaluations~\cite{wang2024mmlu,zheng2023judging}, where open-sourced models generally outperform closed-sourced ones. Specifically, Llama3.1-70B-Instruct outperforms GPT-4o on the inferring skill, and Qwen2-72B-Instruct exceeds GPT-4o on the connecting skill. This inconsistency may result from the differences between comprehension process evaluation and answer-based evaluation. Comprehension process evaluation examines how LLMs arrive at those answers. This evaluation could reveal bottlenecks in the comprehension process, which helps to find ways to improve and advance LLM's comprehension.

\paragraph{Performance w.r.t., document types}
Different document types indeed impact LLMs' performance across various skills. For the locating skill, LLMs perform differently on the SQuAD v2.0 and NewsQA datasets, which share the same answer style but differ in document types. LLMs are better at locating facts in expository documents compared to events in narrative documents. In contrast, for connecting and selecting skills, LLMs excel with narrative texts. This is because comprehending the whole content of a narrative is generally easier than expository documents. These results suggest that LLMs' comprehension is somewhat similar to how humans comprehend documents~\cite{graesser2003readers}.

\paragraph{Performance w.r.t., answer styles}
The comprehension process of LLMs is influenced by answer styles. 
When comparing performance on the MCTest and NewsQA datasets, which have the same document types but different answer styles, we observe that LLMs perform better on multiple-choice questions. This might be because the options provided in these questions reduce cognitive load and make comprehension easier. This phenomenon is similar to how humans perform across different answer styles~\cite{frederiksen1984real}.

\begin{table}[t]
\small 
\begin{tabularx}{\linewidth}{lCCC} 
\toprule
\textbf{Model} & \textbf{Locating}
% \newline\textbf{IC (\%)}
& \textbf{Inferring}
% \newline\textbf{IC (\%)} 
& \textbf{Increase}
% \newline\textbf{(\%)}
\\
\midrule
Qwen2-72B-Instruct & 4.28 & 22.50 & 18.22 \\
Llama3.1-70B-Instruct & 4.45 & 19.68 & 15.23 \\
Claude-3.5-Sonnet & 4.77 & 12.80 & 8.03 \\
GPT-4o & 2.32 & 18.29 & 15.97 \\
\bottomrule
\end{tabularx}
\caption{Inconsistency scores ~(\%) of LLMs on QA data. Scores are averaged across datasets for each skill.}
\label{tab:process_product_analyze}
% \vspace{-5mm}
\end{table}

\begin{figure}[t]
  \centering
  \includegraphics[width=0.88\linewidth]{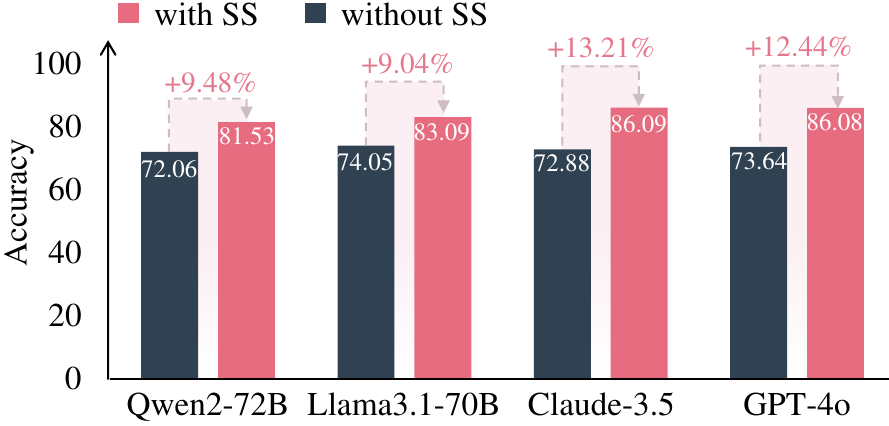}
  \caption{Average accuracy of LLMs with and without supporting sentences~(SS) across the inferring datasets.}
  \label{inferece-ss}
  % \vspace{-5mm}
\end{figure}

\subsection{Auxiliary Analysis}
{We further analyze how the comprehension process affects the answer. The results highlight the importance of a correct comprehension process. Besides, we examine the correlation across five skills, validating the rationality of our framework.}

\paragraph{Correlation between the comprehension process and answers} 
% To explore the relationship between the comprehension process and answers, 
We analyze the correlation between identifying supporting sentences and answering questions at both the locating and inferring levels. 
{We report the inconsistency score, which is defined as the proportion of samples with incorrect supporting sentences and correct answers.}
% \footnote{Accuracy of span-answers is checked using GPT-4o-mini.}.
The results in Table~\ref{tab:process_product_analyze} reveal that LLMs exhibit inconsistent behaviors even at the basic locating level. {This is a strong evidence to support our statements that matching answers alone cannot provide a reliable judgment and more efforts are required in comprehension process evaluation. Moreover, we observe that the inconsistency score tends to increase with the complexity of comprehension process. For example, the inconsistency score of GPT-4o is over 7 times higher at the inferring level than at the locating level. After a close look at the failure samples, we find that LLMs may be ``slacking off'' --- they prefer to use unexplainable shortcuts when they cannot identify all the supporting sentences. This emphasizes the importance of studying the comprehension process for more explainable LLMs.}
% \xiao{This implies that relying solely on answer accuracy may not fully evaluate an LLM's comprehension. It's crucial to focus on process evaluation.
% Moreover, as the comprehension process becomes more complex, inconsistent behaviors worsen. For example, GPT-4o's inconsistency score is over 7 times higher at the inferring level than at the locating level. Through the examination of inconsistent inferring cases, we observe that LLMs would miss some supporting sentences. LLMs with the incomplete comprehension process answer correctly, indicating their unreliability. }
% This might happen due to a lack of intermediate supervision during training, leading to reliance on shortcuts.

\paragraph{How the comprehension process affects answers}
{A simple question is whether the correctness of comprehension process truely makes a difference for LLMs.} This can be investigated at the inferring level, where the golden supporting sentences can construct correct comprehension processes and the predicted answers can used as the observation targets. As shown in Figure~\ref{inferece-ss}, {with the full set of supporting sentences}, the performances of all LLMs significantly increase, suggesting the importance of the comprehension process for LLMs. 
We further notice that the least lazy LLM~(Claude-3.5) gets the biggest benefits~(13.21\%) from the correct comprehension process. Thus, we believe that it will be much easier to achieve expert-level comprehension by adding more supervision to prevent LLMs from using shortcuts during training.}

\begin{figure}[t]
  \centering
  \resizebox{\linewidth}{!}{\includegraphics{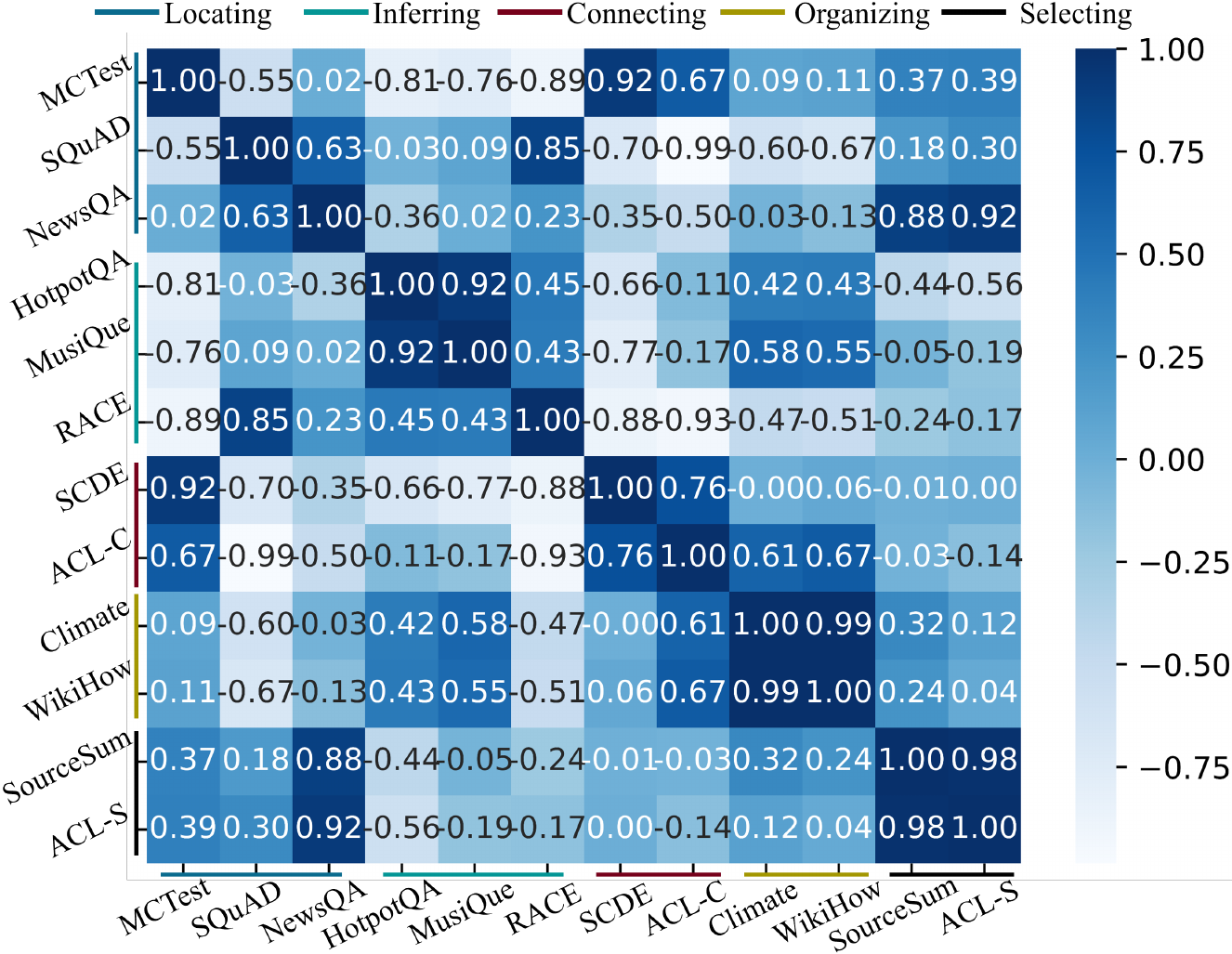}}
  \caption{Inter-task performance correlation.}
  \label{skills_co}
  % \vspace{-5mm}
\end{figure}

\paragraph{Correlation among comprehension process skills}\quad 
Finally, we validate our framework by analyzing the correlations in LLMs performance across all datasets, as shown in Figure~\ref{skills_co}.
Our observations are as follows:
1) Datasets evaluating the same skill generally correlate. MCTest is an exception as it is based on simple children's stories. All LLMs perform similarly on this dataset, leading to a lower correlation with others.
2) There is a correlation between locating and inferring skills since inferring builds on locating. The inferring skill involves both locating and aggregating information in two sub-stages.
3) As expected, all three skills at the interpreting level correlate.
4) Some skills are negatively correlated as they emphasize different aspects.
For instance, while both connecting and selecting involve comprehending the entire document, connecting focuses on relationships between sentences for a broader view, whereas selecting targets key information for a more centralized view. 
5) Tasks at the interpreting level correlate with both locating and inferring skills, indicating that global comprehension relies on the skills required for local comprehension.
These correlations across comprehension process skills align with the expert comprehension process ~\cite{afflerbach2015conceptualizing}, confirming the effectiveness of our framework.

\section{Related Work}
\paragraph{Comprehension Evaluation}
Towards the application of LLMs in real-world scenarios, it is necessary to ensure LLMs share a similar comprehension process with experts. However, current comprehension evaluation methods either focus on the stage after the process~(matching the answers with human references)~\cite{rajpurkar-etal-2018-know,lai-etal-2017-race,dua-etal-2019-drop} or the stage before the process~(digging the literary skills of LLMs)~\cite{dunietz-etal-2020-test,sugawara-etal-2017-evaluation,wang-etal-2022-feeding,sugawara-etal-2021-benchmarking}. To this end, we propose \name to carefully examine the gap between the comprehension processes of LLMs and experts.

\paragraph{Design of Comprehension Tasks}
Recent years have witnessed plenty of emerging reading comprehension tasks, exploring the comprehension potential of LLMs from different aspects. Some studies focus on challenging LLMs with different document types~\cite{kovcisky2018narrativeqa, dasigi2021dataset}. 
% different document type: hudson2022muld
They argue that different types of documents need different ways of comprehension, as narrative documents
% ~(e.g., news, fairy tales) 
mostly describe events while expository documents
% ~(e.g., technical articles, encyclopedias) 
mostly explain facts. In line with these studies, we also involve both narrative and expository documents in \name. Another part of the literature works on the surface forms of the tasks, including text extraction~\cite{rajpurkar-etal-2018-know, dasigi2019quoref}, multi-choice answers~\cite{richardson-etal-2013-mctest, clark2018think}, free-from answers~\cite{bajaj2016ms}, etc. For a full and objective evaluation, we also involve diverse styles of answers but exclude free-form ones, the correctness of whose answers is a matter of preference.
There are also studies working on adding ``difficulties'' to the comprehension burdens. They require LLMs to do causal reasoning~\cite{jin2024cladder}, commonsense reasoning~\cite{zhao2024large}, arithmetic calculation~\cite{yuan2023well}, etc. We think this is beyond the comprehension process thus we don't involve such studies in \name.

\section{Conclusion}
% The demand for reliable AI applications necessitates a thorough analysis of LLMs' comprehension process. 
We propose \name to explore the comprehension process of LLMs from a cognitive view. 
% \replace{We systematically define the skills required in the comprehension process, equipping each skill with a well-matched task and strict screening testing data.}
Specifically, it is equipped with a systematical definition of five requisite comprehension skills, a strict data construction framework, and a detailed analysis of various LLMs.
Experimental results reveal that all LLMs are still far from achieving the expert-level comprehension process. 
% \remove{We further analyze the correlation between the comprehension process and question-answering in LLMs.} 
Further analysis reveals LLMs exhibit inconsistent behavior. They use incorrect comprehension processes despite providing correct answers, highlighting the importance of comprehension processes for reliable models. 
Additionally, we observe that a better comprehension process can lead to better downstream performance.
We hope \name can benefit the research on the improvement and deployment of reliable LLMs. 
% In future work, we will try to introduce comprehension skills during the training of LLMs for improvement. 
% Moreover, using internal knowledge during the comprehension process could be valuable in real-world applications.
\section{Ethic Consideration and Limitations}
The testing data are constructed from both existing and newly crawled datasets. For existing datasets, they are all accessed and used in full compliance with their respective licenses and terms of use. Each dataset was reviewed to ensure that the permissible uses under the applicable licenses align with the scope of our research. For newly crawled datasets, we have carefully checked them to make sure they don’t contain any personally identifiable information or sensitive personally identifiable information. Therefore, we believe that there is no ethical issue within \name.
\par
Part of our testing data is from existing datasets, so some data might already be known to LLMs. In the future, it's important to detect previously seen data and develop better data generation methods. Like other studies on prompting LLMs, our evaluation results may be sensitive to the prompts used. 
The five skills interact in complex ways, making it difficult to directly confirm the relationships between them. Fortunately, educational research has explored these relationships in depth~\cite{rampey2009naep}. We believe exploring these in NLP could deepen our understanding of LLMs' comprehension from a cognitive perspective.
We focus only on the comprehension process as they are fundamental. The next step is to study how to combine internal knowledge during the thinking process. We believe such thinking is necessary for a better comprehension of LLMs. 
% We only used four large language models in our \name benchmark due to computational constraints. With additional resources and an improved experimental environment, evaluating the performance of other LLMs would be advantageous, such as Llama-3.1-405B.
% 数据集都是for research purpose, 符合xxx标准，我们的处理主要是过滤和分类，没有引入任何伦理风险，我们相信没问题
% 我们用的现有数据集，不能避免肯定有些已经暴露给了LLM，未来需要设计方法检测哪些数据已经暴露给LLM，还需要研究更好的数据生成方法xxxxx
% 

\section*{Acknowledgement}
This work was supported in part by the National Natural Science Foundation of China (No. 62206191, No. 62272330, No. U24A20328, and No.62306156), in part by the Science Fund for Creative Research Groups of Sichuan Province Natural Science Foundation (No. 2024NSFTD0035), in part by the Natural Science Foundation of Sichuan (No. 2024YFHZ0233), and in part by the Undergraduate Education and Teaching Reform Projects, Nankai University (NKJG2025047).

\bibliography{reference}

\appendix
\appendix

\section{Data Collection Details}
\label{collection_detail}

\subsection{Data Statistics}
More details about the testing data are shown in Table~\ref{stastics_documents}, which includes statistics on document length, the original dataset size, and associated cost.

\begin{table*}[t]
\centering
\scalebox{0.87}{
\begin{tabular}{lcccccc}
\toprule
\textbf{Skill} & \textbf{Source} & \textbf{Mean Tokens} & \textbf{\makecell[c]{Max Tokens}} & \textbf{Original Size}  & \textbf{Size} & \textbf{Cost (dollars)}\\ \hline
\multirow{3}{*}{Locating}   
& SQuAD v2.0  & 187 & 714 & 5928 & 479 (8.08\%) & 0.72
\\ \cline{2-7} 
& NewsQA & 735 & 2463 & 10292 & 984 (9.56\%) & 0.89
\\ \cline{2-7} 
& MCTest & 253 & 514 & 1160 & 72 (6.21\%) & 0.11  
\\ \hline
\multirow{3}{*}{Inferring}  
& HotpotQA & 1329 & 3035  & 7405 & 604 (8.16\%) & 0.69
\\ \cline{2-7} 
& MusiQue & 2413 & 4449 & 2417 & 510 (21.10\%) & 0.40
\\ \cline{2-7} 
& RACE & 338 & 974 & 9821 & 547 (5.57\%) & 1.25
\\ \hline
\multirow{2}{*}{Connecting} 
& SCDE & 276 & 1010 & 625 & 625 (100.00\%)	& -
\\ \cline{2-7} 
& ACL OCL & 514 & 1104 & - & 169 ( - ) & -
\\ \hline
\multirow{2}{*}{Organizing} 
& ClimateCentral & 1046 & 3596 & - & 108 ( - ) & -
\\ \cline{2-7} 
& WikiHow & 1459 & 4583 & - & 146 ( - ) & -
\\ \hline
\multirow{2}{*}{Selecting}  
& SourceSum & 294 & 542 & 143 & 143 (100.00\%) & -
\\ \cline{2-7} 
& ACL OCL & 535 & 1490 & - & 295 ( - ) & -
\\ \bottomrule
\end{tabular}
}
\caption{Statistics on document length, size, and associated cost of the testing data.}
\label{stastics_documents}
\end{table*}

\subsection{Different Document Types}
According to \citet{eason2012reader}, the main types of documents include narrative and expository. Narrative documents are typically structured around events about characters in a temporal sequence. Examples of narrative structures include news articles, stories, and fiction. On the other hand, expository documents focus on presenting facts about a specific topic. Common examples of expository texts are encyclopedias and reports.

\subsection{Semantic Similarity Calculating}
\label{semantic}
To select an effective model for calculating semantic similarity, we use the MTEB benchmark \cite{muennighoff-etal-2023-mteb}. MTEB tests how well text embeddings perform on different tasks. We focus on four tasks from MTEB that are most relevant for finding semantically similar sentences: Clustering (the task of grouping similar documents together), Pair classification (the task of determining whether two texts are similar), Retrieval (the task of finding relevant documents for a query), and STS (the task of determining how similar two texts are). We add up the models' scores across these tasks and pick the one with the highest overall score. We prefer a smaller-scale model(less than 500M parameters) because it often performs better in sentence-pair similarity tasks and reduces computational costs. Finally, we choose stella-en-400M-v5\footnote{\url{https://huggingface.co/dunzhang/stella_en_400M_v5}} for calculating similarity using cosine similarity scores.

\begin{table*}[t] 
\resizebox{\linewidth}{!}{
\begin{tabular}{l|ccc|ccc|cc|cc|cc}
\toprule
\multirow{2}{*}{\textbf{Model}} & \multicolumn{3}{c|}{\textbf{Locating}}                 & \multicolumn{3}{c|}{\textbf{Inferring}}              & \multicolumn{2}{c|}{\textbf{Connecting}} & \multicolumn{2}{c|}{\textbf{Orgnazing}}    & \multicolumn{2}{c}{\textbf{Selecting}} \\ \cline{2-13} 
                                    & \textbf{MCTest} & \textbf{SQuAD} & \textbf{NewsQA} & \textbf{HotpotQA} & \textbf{MusiQue} & \textbf{RACE} & \textbf{ \ SCDE \ }     & \textbf{ACL}     & \textbf{Climate} & \textbf{WikiHow} & \textbf{SourceSum}  & \textbf{ACL} \\ \hline
Deepseek-R1              & 97.22           & 99.16              & 86.89           & 51.32            & 36.67            & 44.97         & 72.16             & 40.83                & 22.22                   & 34.25            & 13.99               & 10.17             \\
GPT-4o-mini                     & 91.67           & 97.49              & 85.87           & 34.44             & 11.37            & 37.84         & 68.25             & 10.06                & 12.04                   & 30.82            & 10.49               & 12.54 
\\ \bottomrule
\end{tabular}}
\caption{Performance of additional LLMs' comprehension process across five skills.} 
\label{app:overallperformance}
% \vspace{-5mm}
\end{table*}

\subsection{Details of Locating Questions Collection}
\begin{figure}[t]
  \centering
  \includegraphics[width=\linewidth]{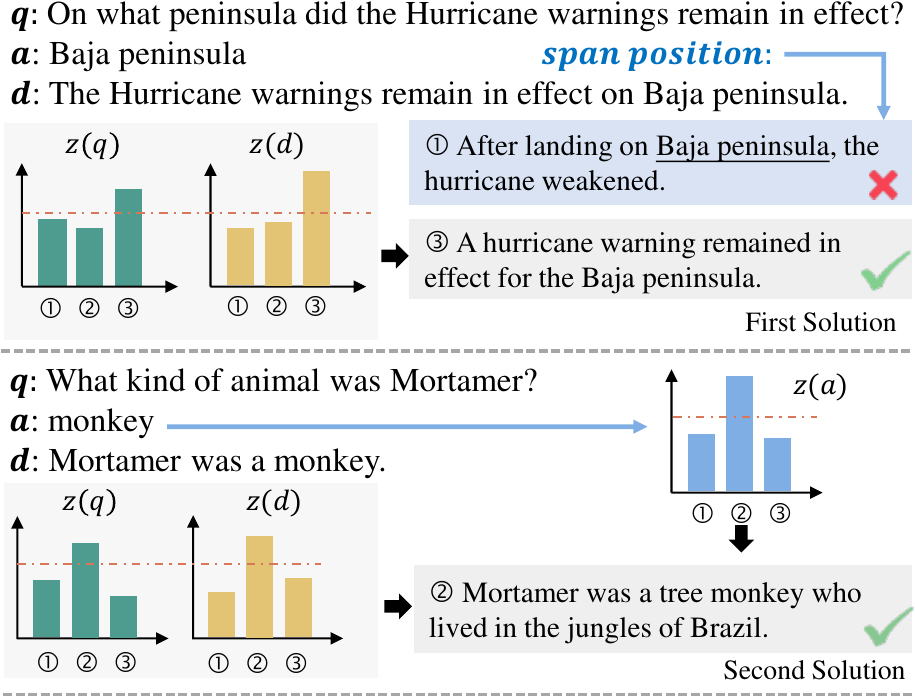}
  \caption{The two solutions for identifying locating-type questions. The first solution is used to handle datasets with answer annotations, while the second handles datasets without them.}
  \label{locating}
\end{figure}

When identifying locating questions, we retain data with the document including more than 4 sentences across the three datasets. As LLMs are required to identify supporting sentence, fewer sentences would impact evaluation results. For MCTest, we also exclude data labeled ``multiple'', meaning a question can be supported by multiple sentences. After preprocessing, we use two solutions described in Section~\ref{sec:loc} (illustrated in Figure~\ref{locating}) to filter locating questions. For the second solution, We observe that $S(\ast, \star)$ doesn't work well when the answer is too short to be well-encoded in embedding spaces. As such, the pseudo-supporting sentence is retrieved by BM25~\cite{robertson1994some}.
Finally, the first solution enables us to filter about 40\% of questions with one supporting sentence in SQuAD 2.0 and over 50\% in NewsQA. The second solution filters about 40\% locating questions in MCTest. 

\paragraph{Impact of the Answer on Retrieval}
When the answer to a question appears only once in the context, using only the declarative sentence for retrieval would be impacted. Taking Table~\ref{answer_impact} as an example, the annotated supporting sentence would be retrieved using the declarative sentence as the query. This question would then be classified as a locating question. However, this annotated supporting sentence does not provide sufficient information to answer the question. To address this issue, we use both the declarative sentence and the question to retrieve. The final set of supporting sentence candidates is the intersection of these retrieval results. This method ensures that the retrieved sentence contain information relevant to both the question and the answer.

\begin{table}[t]
\setlength{\arrayrulewidth}{1pt} 
\begin{tabularx}{\linewidth}{X}
\toprule
\textbf{Document}: In April 1191 Richard the Lion-hearted left Messina with a large fleet in order to reach Acre. But \underline{a storm} dispersed the fleet. After some searching, it was discovered that the boat carrying his sister and his fiancée Berengaria was anchored on the south coast of Cyprus.
\\
\\
\textbf{Question}: What ruined Richard's plans to reach Acre? \\
\textbf{Answer}: a storm
\\
\bottomrule
\end{tabularx}
\caption{An example where the answer appears only once in the context. The annotated span is in bold, and the annotated supporting sentence is underlined. }
\label{answer_impact}
\end{table}

\subsection{Details of Inferring Questions Collection}
We use the pretrained model en\textunderscore core\textunderscore web\textunderscore trf\footnote{\url{https://spacy.io/models/en}} in SpaCy to perform constituency parsing.

% The parsing trees for bridge questions and comparison questions are shown in Figure~\ref{bridge} and Figure~\ref{comparison}, respectively. These figures also show the process of pruning subquestion candidates described in Section~\ref{sec:inf}.
% \begin{figure}[t]
%   \centering
%   \includegraphics[width=\linewidth]{Figure/bridge_infer.pdf}
%   \caption{Illustrations of a non-inferring question~(upper) and inferring question~(lower)}
%   \label{bridge}
% \end{figure}

% \begin{figure}[t]
%   \centering
%   \includegraphics[width=\linewidth]{Figure/com.png}
%   \caption{Illustrations of a non-inferring question~(upper) and inferring question~(lower), presenting the comparison structure.}
%   \label{comparison}
% \end{figure}
% \paragraph{Additional Filtering Process for Subquestions}
Some subquestion candidates cannot be subquestions even if they are noun phrases with named entities. For example, in the question ``Who founded the company that distributed the film UHF?'', the candidate ``the film UHF'' is not a subquestion because it does not target anything. In contrast, in the question ``Who is the partner of Green performer?'', the candidate ``the Green performer'' is a valid subquestion becuase of targeting ``Who is the performer of the Green''.
These candidates are always found at the deepest edges of the syntactic trees. We check their validity by turning them into yes-no questions. For example, ``the film UHF'' is transformed into  ``Is UHF the film'' and ``the Green performer'' is transformed into ``Is Green performer?'' through template filling. We then ask Llama-3-8B-Instruct to answer these questions with the corresponding documents. If the answer is ``No'', this candidate is confirmed as a subquestion. 

After filtering out invalid candidates, we remove those candidates targeting the same fact. We use kendall's tau coefficient to measure the correlation between two similarity distribution. This coefficient is useful for data considering rankings, which helps us focus on the similarity rankings between candidates and document sentences. The part-of-speech used to identify inferring questions from comparison questions is performed by SpaCy.

\subsection{Details of Connecting Data Collection}
Noisy papers with messy code from ACL OCL corpus are filtered using regular expression. Besides, we only retain papers published at EMNLP and ACL conferences because these papers have a unified structure. This ensures data consistency.

\subsection{Details of Organizing Data Collection}
We use the Python library Scrapy\footnote{\url{https://scrapy.org/}} to develop a crawler for extracting documents from the ClimateCentral website. We retain documents with more than two subheadings, ensuring that LLMs would need to segment the documents into at least three subsections. For the Wikihow dataset, we filter out documents that contain too many short sentences (fewer than 5 words), as such documents are more like procedural text rather than expository text.

\subsection{Details of Human Evaluation}
To assess the validity of our testing data, we randomly pick 50 samples from each dataset. We instruct three workers with English level certificates to label whether the given sample meets the definition of the specific tasks. If all workers agree that the sample is valid, it is scored 1, otherwise 0. The average validity score is calculated by summing these scores and dividing by the total number of samples. As a result, the validity score is 0.81 with an inter-annotator agreement percentage of 73\%, indicating a high reliability of our testing data.

\section{Additional Experiments}
\label{Memorized_exp}
\subsection{The Influence of Memorized Data on Performance}
In this pilot experiment, we explore how memorized data affects the evaluation performance of LLM. We classify questions that can be directly answered by an LLM without any context as ``memorized data'', while questions requiring reasoning or additional context are categorized as ``non-memorized data''. Using this distinction, we evaluate the performance of Llama 3.1-70B on both types of data. The results, shown in Table~\ref{memorized}, indicate that the model performs better on memorized data compared to non-memorized data. To reduce the influence of data contamination on our evaluation, we filter out memorized data in Section~\ref{quality}.
\begin{table}[h]
\centering
\scalebox{0.95}{
\begin{tabular}{lccc}
\hline
 & HotpotQA & MusiQue & RACE \\
\hline
non-memorized & 62.58 & 30.98 & 46.07 \\
memorized & 62.75 & 32.93 & 49.17 \\
\hline
\end{tabular}}
\caption{Performance of Llama3.1-70B on memorized vs. non-memorized data.}
\label{memorized}
\end{table}

\subsection{Exploring Additional LLMs with \name}
\label{app:overll}
In Table~\ref{app:overallperformance}, we observe Deepseek-R1 and GPT-4o-mini perform consistently with our findings obtained from \name. Deepseek-R1, with its strong reasoning capability,  outperforms most LLMs in the inferring skill. In a pilot study, we also test Llama-3.1-7B but exclude it from the main evaluation due to its extremely poor performance (e.g., 0.59\% in connecting, 7.4\% in organizing, and 3.4\% in selecting).

\subsection{Case Study}
\label{app:case}
We present a failure example from the climate dataset for the organizing skill using GPT-4o, shown in Table~\ref{case}. This task requires LLMs to determine where subheadings should be placed within a document. In this case,  the correct position for the subheading is clearly (28). However, GPT-4o incorrectly place it at (29). This mistake may be due to the overlapping token "measure" in both the subheading and sentence (29). This indicates LLMs may not follow a correct comprehension process but rather rely on shortcuts. 

\begin{table}[t]
\setlength{\arrayrulewidth}{1pt} 
\begin{tabularx}{\linewidth}{X}
\toprule
\textbf{Document}: \\
...(24) There are efforts underway to similarly capture and use methane from agriculture and waste facilities known as biogas. (26) There is a growing range of strategies to otherwise reduce methane production from farming and landfills, including alternative feed for cows and composting to reduce waste. (27) Climate Central’s full report on methane elaborates on sector-specific strategies and key initiatives to reduce methane from these main sources.\\
(28) Accurate emissions information is key for setting priorities and tracking progress–or lack thereof–toward reduction goals. (29) But collecting data on an odorless, invisible gas requires specialized technology, so direct measurements of methane emissions are thin. 
...\\
\\
\textbf{Subheading}: How methane is measured \\
\bottomrule
\end{tabularx}
\caption{A failure case where GPT-4o fails to organize. We add a line break at sentence (28) for clarity. The original document does not have this break.}
\label{case}
\end{table}

We use an example from the RACE dataset to explain why we do not directly treat the sentences retrieved by LLM as the golden supporting sentences. Using the sample Table~\ref{case2} as an example, the LLM selects sentences 8, 11, 12, and 13 as the supporting sentences. However, treating these sentences as the golden results and using this sample as an inference question conflicts with our definition of the comprehension process.
This is because answering the question requires knowledge about human emotions: a person feels happy when they achieve their goal. This kind of knowledge is not explicitly stated in the document. 
To address this, we also extract supporting sentences based on semantic similarity, resulting in sentences 7, 8, and 9. Since fewer than two of these sentences overlap with the ones chosen by the LLM, we exclude this sample from the inference questions. This case shows that the supporting sentences extracted by LLMs are influenced by their background knowledge. Therefore, we incorporate semantic similarity to ensure that the test data more accurately matches the required definition.

\begin{table}[t]
\setlength{\arrayrulewidth}{1pt} 
\begin{tabularx}{\linewidth}{X}
\toprule
\textbf{Document sentences}: \\
(1) Long, long ago there was an old man. \\
(2) He had a very big orange tree in his garden. \\
(3) On the tree there were many fine oranges. \\
(4) One day the old man found one of the oranges was bigger than the others. \\
(5) It was as big as a watermelon. \\
(6) So he took the big orange to the king. \\
(7) The king was very happy and gave the old man a lot of money for it. \\
(8) When a rich man heard of this, he said to himself, "It is only an orange." \\
(9) "Why did the king give him so much money?", \\
(10) "If I take my gold cup to the king, he will give me much more money for it.", \\
(11) The next day when the king got the gold cup, he said to the rich man, "What a beautiful cup!" \\
(12) "I'll give you something for it."\\
(13) "Please take the great orange."\\
\\
\textbf{Questions}: Was the rich man very happy at last? \\
\bottomrule
\end{tabularx}
\caption{A sample from RACE.}
\label{case2}
\end{table}

\clearpage
\onecolumn
\section{Data Examples}
\label{data_examples}
\subsection{Locating}

\begin{table*}[h]
\small
\setlength{\arrayrulewidth}{1pt} 
\begin{tabularx}{\textwidth}{X}
\toprule
\textbf{Document}: One of the first Norman mercenaries to serve as a Byzantine general was Hervé in the 1050s. By then however, there were already Norman mercenaries serving as far away as Trebizond and Georgia. They were based at Malatya and Edessa, under the Byzantine duke of Antioch, Isaac Komnenos. In the 1060s, Robert Crispin led the Normans of Edessa against the Turks. Roussel de Bailleul even tried to carve out an independent state in Asia Minor with support from the local population, but he was stopped by the Byzantine general Alexius Komnenos.\\
\\
\textbf{Question}: When did Hervé serve as a Byzantine general? \\
\textbf{Answer}: in the 1050s \\
\textbf{Supporting sentence}: One of the first Norman mercenaries to serve as a Byzantine general was Hervé in the 1050s. \\
\bottomrule
\end{tabularx}
\caption{Example of a locating question in SQuAD 2.0}
\label{squad}
\end{table*}
% \vspace{-7mm}

\begin{table*}[h]
\small
\setlength{\arrayrulewidth}{1pt} 
\begin{tabularx}{\textwidth}{X}
\toprule
\textbf{Document}: Authorities have recovered 54 bodies after a ferry crammed with people capsized in southern Bangladesh , police said Sunday . Among the victims were 22 children and 15 women , said Nazrul Islam , the police chief of Bhola district where the accident occurred Friday . Thirty more passengers are believed missing and presumed dead , he said . `` Hopefully , in few hours , we should be able to confirm the exact number of missing -LRB- people -RRB- , '' Islam said . The boat had a capacity of 1,500 but was overcrowded with about 2,000 people who were traveling from the capital , Dhaka , to their homes in Bhola for the Muslim festival of Eid al-Adha . The boat toppled as passengers weighted down one side to disembark , Islam said . Police and firefighters rushed to aid passengers , many of whom were trapped in the lower deck . CNN 's Harmeet Shah Singh contributed to this report. \\
\\
\textbf{Question}: what was traveling ? \\
\textbf{Answer}: 2,000 people \\
\textbf{Supporting sentence}: The boat had a capacity of 1,500 but was overcrowded with about 2,000 people who were traveling from the capital , Dhaka , to their homes in Bhola for the Muslim festival of Eid al-Adha . \\
\bottomrule
\end{tabularx}
\caption{Example of a locating question in NewsQA}
\label{newsqa}
\end{table*}

\begin{table*}[h!]
\small
\setlength{\arrayrulewidth}{1pt} 
\begin{tabularx}{\textwidth}{X}
\toprule
\textbf{Document}: Jenny was standing on a rock.  Suddenly, she had to sneeze.  After she sneezed, she walked away.  She finally got to the park and saw her daddy.  Her daddy gave her some milk.  Jenny drank the milk in a big hurry.  She loved milk.  She walked over and turned a switch.  She walked to the lake.  Jenny was in a big hurry and went really fast.  She got to the lake and sat down.  Jenny began thinking.  Jenny wanted to go on a trip to Florida.  Jenny did not want to go someplace cold.  Jenny did not want to go to the moon.  Jenny did not want to go to France.  Jenny stood up to fold her towel.  She never folded her shirts or pants.  Jenny would start her art for her aunt in a few hours.  She knew she would use a lot of time making that art.  Her aunt would love the art. \\
\\
\textbf{Question}: Where did Jenny want to go on a trip to? \\
\textbf{Options}: A. Florida \qquad B. someplace cold \qquad C. France \qquad D. the moon \\
\textbf{Answer}: A \\
\textbf{Supporting sentence}: Jenny wanted to go on a trip to Florida. \\
\bottomrule
\end{tabularx}
\caption{Example of a locating question in MCTest}
\label{mctest}
\end{table*}

\clearpage

\subsection{Inferring}
\begin{table*}[h]
\small
\label{race}
\setlength{\arrayrulewidth}{1pt} 
\begin{tabularx}{\textwidth}{X}
\toprule
\textbf{Document}: \\
I work at a university in the USA. There, my team and I are trying to learn more about the American black duck, a kind of water bird. And now we are using an exciting piece of equipment called a ``night vision scope''. By using it, we can see the ducks in the dark. We're worried about black ducks mainly because their numbers are falling \underline{\hspace{2cm}}. And we don't know whether there's enough food on the east coast for these birds. There's lots of information about their daytime activities, but nothing about what they do at night, because we don't have the equipment. But this new ``scope'' will make really clear pictures, even on moonless nights, so we will be able to find out more about the ducks.It is very hard work. There are four of us, and we each work six hours every day. We study ducks in different places, and I sometimes have to take a boat to where I need to work. The weather is not helpful because most of the time it's wet... \\
\\
\textbf{Question}: What does the writer' team hope to find out about American black ducks? \\
\textbf{Options}: 
\\ A. What food they feed on. \qquad B. What makes the east coast a good place for them. \\
C. What they do at night. \qquad  D. What animals like to stay with them. \\
\textbf{Answer}: C \\
\textbf{Supporting sentence}: \\
1. There's lots of information about their daytime activities, but nothing about what they do at night,because we don't have the equipment. \\
2. But this new ``scope'' will make really clear pictures, even on moonless nights, so we will be able to find out more about the ducks. \\
\bottomrule
\end{tabularx}
\caption{Example of an inferring question in RACE}
\end{table*}

\begin{table*}[h!]
\small
\label{hotpotqa}
\setlength{\arrayrulewidth}{1pt} 
\begin{tabularx}{\textwidth}{X}
\toprule
\textbf{Document}: 
\\
Klaus Meine || Klaus Meine (born 25 May 1948) is a German vocalist, best known as the lead singer of the hard rock band Scorpions. He and guitarist Rudolf Schenker are the only two members of the group to appear on every Scorpions album. Meine was placed at \#22 on Hit Parader's Top Heavy Metal Vocalists of All Time list in 2006. \\
A Moment in Chiros || A Moment in Chiros is American heavy metal vocalist Lance King's studio debut album as a solo artist, featuring the musical contributions of many of his friends, contemporaries, and business associates. \\
Geoff Tate || Geoff Tate (born Jeffrey Wayne Tate, January 14, 1959; he later changed his first name to Geoffrey) is a German-born American singer and musician. He rose to fame with the progressive metal band Queensrÿche, who had commercial success with their 1988 album ``Operation: Mindcrime'' and 1990 album ``Empire''. Tate is ranked fourteenth on Hit Parader's list of the 100 Greatest Metal Vocalists of All Time. He was voted No. 2 on ``That Metal Show''\textquotesingle s top 5 hard rock vocalists of the 1980s. In 2012, he won the Vegas Rocks! Magazine Music Award for ``Voice in Progressive Heavy Metal''. In 2015, he placed ninth on OC Weekly's list of the 10 Best High-Pitched Metal Singers. After his farewell tour as Queensrÿche, he renamed his band Operation: Mindcrime, after the Queensrÿche album. \\
Dee Snider || Daniel ``Dee'' Snider (born March 15, 1955) is an American singer-songwriter, screenwriter, radio personality, and actor. Snider came to prominence in the early 1980s as lead singer of the heavy metal band Twisted Sister. He was ranked 83 in Hit Parader\textquotesingle s Top 100 Metal Vocalists of All Time. \\
Lance King || Lance King (born November 23, 1962) is an American heavy metal vocalist specializing in melodic rock, progressive, and power metal. Lance has sung with many groups over the last 35 years and started the record label Nightmare in 1990 to release his own music. He is presently still at the helm of the label. \\
Avian (band) || Avian is a melodic power metal band founded in 2002 by guitarist Yan Leviathan. The band features singer Lance King. In 2005, they released their debut album ``From the Depths of Time'', a concept album dealing with the end of days and a warning to mankind. Musically, Avian is influenced by bands such as Iron Maiden, HammerFall, Savatage, and Megadeth. In December 2006, Avian was an opening act for Twisted Sister. Their second album, titled ``Ashes and Madness'', was released in September 2008. In early 2010, Lance decided to leave the band to focus on family and professional obligations and was replaced with Brian Hollenbeck, who appeared on their first EP, entitled ``The Path'', which was released in September 2010. \\
Sully Erna || Salvatore Paul ``Sully'' Erna Jr. (born February 7, 1968) is the American vocalist and guitarist for the American hard rock band Godsmack. He is also a harmonica player, percussionist, and pianist, performing these on albums and at live shows. He was ranked 47th in the Top 100 Heavy Metal Vocalists by Hit Parader. \\
Han Seung-yeon || Han Seung-yeon (born July 24, 1988), better known mononymously as Seungyeon, is a South Korean singer and actress. She is best known as the former main vocalist of the South Korean girl group Kara.\\
...\\
\\
\textbf{Question}: Are Lance King and Han Seung-yeon both heavy metal vocalists? \\
\textbf{Answer}: No \\
\textbf{Supporting sentence}: \\
1. Lance King (born November 23, 1962) is an American heavy metal vocalist specializing in melodic rock progressive and power metal.\\
2. She is best known as former main vocalist of the South Korean girl group Kara. \\
\bottomrule
\end{tabularx}
\caption{Example of an inferring question in HotpotQA}
\end{table*}

\clearpage

\begin{table*}[h]
\small
\setlength{\arrayrulewidth}{1pt} 
\begin{tabularx}{\textwidth}{X}
\toprule
\textbf{Document}: 
\\
West DeLand, Florida || West DeLand is a census-designated place (CDP) in Volusia County, Florida, United States. The population was 3,535 at the 2010 census. \\
Kendall Green, Pompano Beach, Florida || Kendall Green was a census-designated place (CDP) in Broward County, Florida, United States, and is now a neighborhood of Pompano Beach, Florida. The population was 3,084 at the 2000 census. \\
Ridgecrest, Florida || Ridgecrest is a census-designated place (CDP) in Pinellas County, Florida, United States. The population was 2,558 at the 2010 census. \\
Wade Hampton, South Carolina || Wade Hampton is a census-designated place (CDP) in Greenville County, South Carolina, United States. The population was 20,622 at the 2010 census. It is named for American Civil War general and South Carolina governor Wade Hampton. \\
Zephyrhills North, Florida || Zephyrhills North is a census-designated place (CDP) in Pasco County, Florida, United States. The population was 2,544 at the 2000 census. \\
Tamiami, Florida || Tamiami is a census-designated place (CDP) in Miami-Dade County, Florida, United States. The population was 55,271 at the 2010 census. \\
Hampton Double Square Historic District || The Hampton Double Square Historic District is a historic district located in Hampton, Iowa, United States. It has been listed on the National Register of Historic Places since 2003. At the time of its nomination it contained 43 resources, which included 28 contributing buildings, two contributing sites, 10 non-contributing buildings, one non-contributing site, one non-contributing structures, and one non-contributing object. The town of Hampton was laid out by H.P. Allen, who was the county surveyor, in June 1856. The original plat was eight blocks by eight blocks in the shape of an ``L''. Near the center of the ``L'' was the two-block, or double, square. While many county seats in Iowa have a courthouse square, the double square is a rarity. Four double squares were platted in Iowa, but only those in Hampton and Sidney survived their early period of development. Estherville's square was platted as a four-block square, but its development created a double square instead. Hampton has the only symmetrical double square plan in the state. The double square exemplifies the two primary functions of a public square, both commercial and public development. \\
Sean Hampton || Born in Ocala, Florida, Hampton is the youngest of five children of a dentist father and a professional model mother. After graduating high school, Hampton enrolled at Stetson University to pursue a career in law. While in school he not only joined Sigma Nu fraternity (Delta Mu chapter), but caught onto acting. After college he married his current wife Jennifer and the two moved to Los Angeles where they currently reside. \\
Gladeview, Florida || Gladeview is a census-designated place (CDP) in Miami-Dade County, Florida, United States. The population was 11,535 at the 2010 census. \\
Solana, Florida || Solana is an unincorporated community and census-designated place (CDP) in Charlotte County, Florida, United States. The population was 742 at the 2010 census. It is part of the Punta Gorda, Florida Metropolitan Statistical Area. \\
Ocala, Florida || Ocala is a city located in Northern Florida. As of the 2013 census, its population, estimated by the United States Census Bureau, was 57,468, making it the 45th most populated city in Florida. \\
... \\
\\
\textbf{Question}: Where is Sean Hampton's birth place in the state of Florida? \\
\textbf{Answer}: in Northern Florida \\
\textbf{Supporting sentence}: \\
1. Born in Ocala, Florida, Hampton is the youngest of five children of a dentist father and a professional model mother.\\
2. Ocala is a city located in Northern Florida. As of the 2013 census, its population, estimated by the United States Census Bureau, was 57,468, making it the 45th most populated city in Florida. \\
\bottomrule
\end{tabularx}
\caption{Example of an inferring question in MusiQue}
\label{musique}
\end{table*}

\subsection{Connecting}
\begin{table*}[h!]
\small
\setlength{\arrayrulewidth}{1pt} 
\begin{tabularx}{\textwidth}{X}
\toprule
\textbf{Document}: Confidence is very important in daily life . It can help you to develop a healthy attitude . But how to be more confident ? Here are some suggestions : <1> If you like singing , sing as much as you can . In some ways , a hobby can make you outstanding . And it will make you happy and confident . <2> Exercise makes you tired but relaxed . A strong body helps you be full of confidence . <3> Fear comes along with failure . But it 's easy to overcome if you know that failure is part of your life . Try to start again and believe you can do better . <4> When you are not confident , you will speak in a low voice . Try to speak loudly enough so that people can hear you clearly . The high voice can help you become more confident . <5> Write down a list of things you did during the day to see how many things you have done well . 
% Did you finish your homework ? Did you tell a joke that made everybody laugh ? Give yourself praise for the good things you' ve done .
\\
\textbf{Candidates}:  \\
1. Play sports . \\
2. Pick up a hobby . \\
3. Speak loudly . \\
4. Get rid of fear . \\
5. Ask for help .\\
6. Find your advantages \\ 
\textbf{Answer}: 2, 1, 4, 3, 6 \\
\bottomrule
\end{tabularx}
\caption{Example of a connecting data in SCDE}
\label{scde}
\end{table*}

\clearpage

\begin{table*}[h!]
\small
\setlength{\arrayrulewidth}{1pt} 
\begin{tabularx}{\textwidth}{X}
\toprule
\textbf{Document}: Text segmentation is a traditional NLP task that breaks up text into constituents, according to predeﬁned requirements. It can be applied to documents, in which case the objective is to create logically coherent sub-document units. <1> This task is often referred to as  document segmentation  or sometimes simply text segmentation . <2> Documents are often multi-modal, in that they cover multiple aspects and topics; breaking a document into uni-modal segments can help improve and/or speed up down stream applications. For example, document segmentation has been shown to improve information retrieval by indexing sub-document units instead of full documents. Other applications such as summarization and information extraction can also beneﬁt from text segmentation <3> breaks up pieces of text into sub-sentence elements called Elementary Discourse Units ( EDUs ). EDUs are the minimal units in discourse analysis according to the Rhetorical Structure Theory. In Figure  2  we show examples of EDU segmentations of sentences. For example, the sentence “Annuities are rarely a good idea at the age 35 because of withdrawal restrictions” decom-poses into the following two EDUs: “Annuities are rarely a good idea at the age 35” and “because of withdrawal restrictions”, the ﬁrst one being a state-ment and the second one being a justiﬁcation in the discourse analysis. In addition to being a key step in discourse analysis, discourse segmentation has been shown to improve a number of downstream tasks, such as text summarization, by helping to identify ﬁne-grained sub-sentence units that may have different levels of importance when creating a summary. <4>  Koshorek et al.  proposed the use of 4708 hierarchical Bi-LSTMs for document segmentation. Simultaneously,  Li et al. introduced an attention-based model for both document seg-mentation and discourse segmentation, and  Wang et al. obtained state of the art results on dis-course segmentation using pretrained contextual embeddings. Also, a new large-scale dataset for document segmentation based on Wikipedia was introduced by  Koshorek et al., providing a much more realistic setup for evaluation than the previously used small scale and often synthetic datasets such as the Choi dataset. However, these approaches are evaluated on different datasets and as such have not been compared against one another. Furthermore they mostly rely on RNNs instead of the more recent transformers and in most cases do not make use of contextual embeddings which have been shown to help in many classical NLP tasks. <5> 1.  We compare recent approaches that were pro-posed independently for text and/or discourse segmentation on three public datasets. 2.  We introduce three new model architectures based on transformers and BERT-style con-textual embeddings to the document and dis-course segmentation tasks. We analyze the strengths and weaknesses of each architecture and establish a new state-of-the-art. 3.  We show that a simple paradigm argued for by some of the earliest text segmentation algorithms can achieve competitive performance in the current neural era. 4.  We conduct ablation studies analyzing the im-portance of context size and model size.\\
\textbf{Candidates}:  \\
1. Our experiments showed that all of our models improve the current state-of-the-art. \\
2. These units, or segments, can be any structure of interest, such as paragraphs or sections. \\
3. In this work we aim at addressing these limitations and make the following contributions: \\
4. A related task called  discourse segmentation \\ 
5. Naturally these results do not imply that hierarchical models should be disregarded. \\
6. In Figure  1  we show one ex-ample of document segmentation from Wikipedia, on which the task is typically evaluated \\
7. Multiple neural approaches have been recently proposed for document and discourse segmentation. \\ 
\textbf{Answer}: 2, 6, 4, 7, 3 \\
\bottomrule
\end{tabularx}
\caption{Example of a locating question in ACL OCL}
\label{acl_connecting}
\end{table*}

\subsection{Organizing}
% \vspace{-3mm}
\begin{table*}[h!]
\small
\setlength{\arrayrulewidth}{1pt} 
\begin{tabularx}{\textwidth}{XX}
\toprule
\multicolumn{2}{>{\hsize=\dimexpr2\hsize+2\tabcolsep+\arrayrulewidth\relax}X}{\textbf{Document}: 
(0) This summer’s relentless record heat has stuck around into fall. (1) The planet just had a record-shattering September—the seventh-warmest for the U.S. (2) This is bad news for the 50 million people in the U.S. with allergies to ragweed pollen in the late summer and early fall. (3) In most U.S. areas, ragweed pollen typically peaks in September and lasts through October. (4) But warmer fall temperatures extend the ragweed growing season. (5) Ragweed, which is found in most U.S. states, is the main cause of fall allergies. (6) A single ragweed plant can produce up to 1 billion pollen grains that are carried by wind and cause a range of symptoms. (7) Ragweed can also thrive in both rural and urban areas. (8) A 2003 study suggests that the urban heat island effect can even help ragweed grow faster and produce more pollen in cities. (9) See Urban Heat Hot Spots to find the strongest urban heat islands within your city. (10) How is fall warming affecting local fall allergy seasons? (11) Climate Central analysis explores this question. (12) Studies have found that the length of ragweed pollen season across the U.S.—from Texas to North Dakota—is strongly linked with the number of fall days until the first frost. (13) Climate Central therefore assessed how the number of consecutive freeze-free days (with minimum temperatures above 32°F) during the fall season (Sept-Nov) has changed since 1970 in 201 U.S. cities. (14) The freeze-free fall season lengthened in 164 cities, or 82\% of the 201 analyzed. (15) Across these 164 cities, the freeze-free fall season lengthened by 11 days on average. (16) The freeze-free fall season is now at least two weeks longer in 53 cities. (17) Over half (58\%) of these 53 cities were in the Northeast, Upper Midwest, and Northwest—consistent with research finding that ragweed pollen season has grown fastest at higher latitudes. (18) The four cities where the fall freeze-free season has grown the most since 1970 are: Reno, Nev. (+39 days); Bend, Ore. (+33 days); Toledo, Ohio (+28 days); Boise, Idaho (+27 days). (19) The widespread increase in freeze-free fall days can prolong allergy-inducing pollen production by the 17 types of ragweed that grow across the U.S. during the late summer and fall. (20) Seasonal allergies can already last from early spring through late fall. (21) But warming from carbon pollution results in more freeze-free days each year, giving plants more time to grow and release allergy-inducing pollen...} \\
\makecell[l]{\textbf{Subheadings:}\\Summer heat lingering into fall\\Growing season lasting later into fall\\Warming climate, longer pollen season, worse allergies\\Mold can cause fall allergies, too}
&
\makecell[X]{\textbf{Answer:}\\Summer heat lingering into fall, 0 \\ Growing season lasting later into fall, 7 \\ Warming climate, longer pollen season, worse allergies, 15 \\  Mold can cause fall allergies, too, 21 } \\
\bottomrule
\end{tabularx}
\caption{Example of an organizing data in ClimateCentral}
% \vspace{-3mm}
\label{climate}
\end{table*}

\begin{table*}[h!]
\small
\setlength{\arrayrulewidth}{1pt} 
\begin{tabularx}{\textwidth}{XX}
\toprule
\multicolumn{2}{>{\hsize=\dimexpr2\hsize+2\tabcolsep+\arrayrulewidth\relax}X}{\textbf{Document}: 
(0) Look for auctions held in less popular or crowded areas. (1) Like any auction, the more crowded it is, the more competition you may have. (2) A big crowd could drive the bidding prices up or cause you to lose out on a bid for a vehicle. (3) Look for auctions that are situated in less populated areas or tend to fly under the radar. (4) You can search for police auctions in certain areas online. (5) Focus on auctions outside of a major city, if possible, or in a smaller town or city, as these may be less crowded than auctions held in larger cities or known areas. (6) Research the vehicles listed online a few days before the auction date. (7) Most auctions will list the vehicles that will be available at the auction a few days before the auction date. (8) Look over each listing and identify which vehicles you are interested in bidding on. (9) You should try to choose at least one to two vehicles in the event you lose out on a bid so you have a backup vehicle you can still bid on. (10) If you have your eye on a Mercedes-Benz CLK listed online, for example, you should note the details listed for the car. (11) Then, you should research the market value of a used Mercedes-Benz CLK and determine how much you would be willing to bid for the car. (12) Make sure you are clear on the maximum you would be willing to spend on the car as this can prevent you from overbidding in the chaos of the auction. (13) Bring cash or proof of an approved loan to the auction. (14) Police auctions will only take payment in cash or proof of an approved loan for the winning bid. (15) If you are planning to pay with an approved loan from your bank, you will need to be able to cover a minimum deposit for the full cost of the vehicle. (16) You will also need to cover the cost of taxes, title, and registration fees. (17) Cars sold at auction do not come with a warranty and are considered ``as is'' so you will likely need to purchase insurance and a warranty for the car once you buy it. (18) You will also need enough money to cover the cost of towing the car from the auction and the cost of cutting new keys for the vehicle if it is sold without keys. (19) Take a set of tools, car oil, and an air pressure gauge. (20) You will not be able to drive the vehicles before you bid on them so inspecting the car beforehand with tools, car oil, and an air pressure gauge can help to ensure the car is in working order. (21) Show up early and check in. (22) The vehicles at a police auction are often shown in a set order so get to the auction early and check in with the auction. (23) You can get a copy of the showing list at check in and have a chance to inspect the vehicles you are interested in before the auction starts. (24) Inspect the vehicles you are interested in bidding on. (25) Use your set of tools to do a quick inspection of the vehicles you plan to bid on. (26) The vehicles appear at the auction untouched, which means they are in the exact state they were in when they were confiscated by the police. (27) Be prepared for the vehicles to be filthy, damaged, or full of someone else’s stuff. (28) Do not be put off by surface level dirt or strong smells, as these can be cleaned out as long the vehicle’s parts are in good shape. (29) Lift the hood of the vehicle and give it a good inspection. (30) Look at the brakes, the shocks, and the quality of the tires on the vehicle. (31) This will help you determine if the vehicle is worth bidding on and how much you should bid for the vehicle. (32) Do not bid more than you can afford. (33) It can be easy to get caught up in the chaos of bidding wars and quick sales at the auction, so focus on staying calm and not bidding more than you can afford. (34) Remember your predetermined limit you set for yourself as you bid on the vehicles you are interested in and try not to overbid in an attempt to outbid someone else. (35) Avoid making quick, in the heat of the moment decisions and really be certain you want a vehicle before you start bidding on it. (36) You don’t want to end up having to pay more for a vehicle than you can afford or than it’s worth because you got caught up in a bidding war. (37) Check if there is a towing company on site. (38)...} \\
\makecell[l]{\textbf{Subheadings:}\\Summer heat lingering into fall\\Attending the Police Impound Auction\\Taking Your Car Home}
&
\makecell[X]{\textbf{Answer:}\\Summer heat lingering into fall, 0 \\ Attending the Police Impound Auction, 21 \\ Taking Your Car Home, 38 } \\
\bottomrule
\end{tabularx}
\caption{Example of an organizing data in Wikihow}
\label{wikihow}
\end{table*}

\subsection{Selecting}
\begin{table*}[h!]
\small
\setlength{\arrayrulewidth}{1pt} 
\begin{tabularx}{\textwidth}{X}
\toprule
\textbf{Document}: Not everyone should be behind the wheel of a \$50,000 car. That's one lesson to take away from a video posted by YouTube user Richard Stewart showing a Porsche Cayman flying out of control as it speeds from a green light on Prince Edward Island in Canada. The sports car swerves wildly before smashing into the concrete median. A wheel even comes off before the car finally comes to a halt. ``Just cause you have a nice car doesn't make you a good driver. Don't let your son drive your Porsche!'' wrote Stewart on YouTube about the crash. KHOU reports that police have not made the identity of the driver public but have said that a 31-year-old driver was cited for the crash, leaving the car totaled as it was towed away. The footage begins with the Porsche idling at a green light. The car booms ahead at a dangerous speed. Almost immediately the driver begins to lose control. The unidentified man veers wildly across the dividing line. The car is twisting at such dangerous speeds a wheel comes loose. Finally, the car comes to a halt, a total wreck waiting for the tow truck. \\
\textbf{Key sentences}:  \\
1. That's one lesson to take away from a video posted by YouTube user Richard Stewart showing a Porsche Cayman flying out of control as it speeds from a green light on Prince Edward Island in Canada. \\
2. KHOU reports that police have not made the identity of the driver public but have said that a 31-year-old driver was cited for the crash, leaving the car a totaled as it was towed away. \\
3. Finally the car comes to a halt, a total wreck waiting for the tow truck . \\
\bottomrule
\end{tabularx}
\caption{Example of a selecting data in sourcesum}
\label{scde}
\end{table*}

\clearpage
\begin{table*}[h!]
\small
\setlength{\arrayrulewidth}{1pt} 
\begin{tabularx}{\textwidth}{X}
\toprule
\textbf{Document}: Sentiment is personal; the same sentiment can be expressed in various ways and the same expression might carry distinct polarities across different individuals. Current mainstream solutions of sentiment analysis overlook this fact by focusing on population-level models. However, only one global model is estimated there, and the details of how individual users express diverse opinions cannot be captured. More importantly, existing solutions build static sentiment models on historic data; but the means in which a user expresses his/her opinion is changing over time. To capture temporal dynamics in a user’s opinions with existing solutions, repeated model reconstruction is unavoidable, albeit it is prohibitively expensive. As a result, personalized sentiment analysis requires effective exploitation of users’ own opinionated data and efficient execution of model updates across all users. To address these challenges, we propose to build personalized sentiment classification models via shared model adaptation. Our solution roots in the social psychology theories about humans’ dispositional tendencies. Humans’ behaviors are shaped by social norms, a set of socially shared ``feelings'' and ``display rules'' about how one should feel and express opinions. Intuitively, personalized model adaptations can be considered as a set of related tasks in individual users, which contribute to a shared global model adaptation. In particular, we assume the distinct ways in which users express their opinions can be characterized by a linear classifier’s parameters, i.e., the weights of textual features. Personalized models are thus achieved via a series of linear transformations over a globally shared classifier’s parameters, e.g., shifting and scaling the weight vector. This globally shared classifier itself is obtained via another set of linear transformations over a given base classifier, which can be estimated from an isolated collection beforehand and serves as a prior for shared sentiment classification. The shared global model adaptation makes personalized model estimation no longer independent, such that regularity is formed across individualized learning tasks. We empirically evaluated the proposed solution on two large collections of reviews, i.e., Amazon and Yelp reviews. Extensive experiment results confirm its effectiveness: the proposed method outperformed user-independent classification methods, several state-of-the-art model adaptation methods, and multi-task learning algorithms. \\
\textbf{Key sentences}:  \\
1. As a result, personalized sentiment analysis requires effective exploitation of users’ own opinionated data and efficient execution of model updates across all users. \\
2. To address these challenges, we propose to build personalized sentiment classification models via shared model adaptation. \\
3. The shared global model adaptation makes personalized model estimation no longer independent, such that regularity is formed across individualized learning tasks. \\
4. We empirically evaluated the proposed solution on two large collections of reviews, i.e., Amazon and Yelp reviews. \\
\bottomrule
\end{tabularx}
\caption{Example of a selecting data in ACL OCL}
\label{scde}
\end{table*}
\clearpage

\section{Prompts for five skill evaluation}
\label{prompts}
\subsection{Locating}
\begin{table*}[h!]
\small
\setlength{\arrayrulewidth}{1pt} 
\begin{tabularx}{\textwidth}{lX}
\toprule
\textbf{Type}        &  \textbf{Prompt}\\
\hline
EQA & \makecell[X]{Answer the provided question based on the given context. First, identify the sentence that supports the answer to the question, then output both the supporting sentence and the answer in JSON format, ensure the answer is directly extracted from the original text. Response with the JSON only!  \\
Here is an example: \\
\#Context\\ 
\#\#context\#\#\\ 
\# Question\\ 
When was the Duchy of Normandy founded?\\ 
\# Output\\ 
\{ 
"supporting\_sentence": "The Duchy of Normandy, which began in 911 as a fiefdom, was established by the treaty of Saint-Clair-sur-Epte between King Charles III of West Francia and the famed Viking ruler Rollo, and was situated in the former Frankish kingdom of Neustria.",\\ "answer": "911" \}\\\\
\# Context\\ \#\#context\#\#\\ \# Question\\ \#\#question\#\#\\ \# Output} \\
\hline

MCQA & \makecell[X]{Given a multiple-choice question, select the correct option based on the provided context. To complete the task, first identify the sentence that supports the answer, and then output both the supporting sentence and the chosen option in JSON format. Response with the JSON only!  \\
Here is an example: \\
\# Context\\ 
\#\#context\#\#
\\\# Question\\ What broke the fence?\\ \# Options\\ {[}"A tree.", "A raccoon.", "John.", "The things that were missing from the back yard." {]}\\ \# Output\\ \{ "supporting\_sentence": "A tree, weighted down by the snow, had fallen on the fence on a windy day and broken a section."
,\\ "answer": "A tree."
\}\\\\
\# Context\\ \#\#context\#\#\\ \# Question\\ \#\#question\#\#\\ \# Options\\ \#\#options\#\#\\ \# Output} \\
\bottomrule
\end{tabularx}
\caption{Prompts for locating skill evaluation.}
\label{locating-prompt}
\end{table*}
\clearpage

\subsection{Inferring}
\begin{table*}[h!]
\small
\setlength{\arrayrulewidth}{1pt} 
\begin{tabularx}{\textwidth}{lX}
\toprule
\textbf{Type}        &  \textbf{Prompt}\\
\hline
EQA & \makecell[X]{Answer the provided question based on the given context. First, identify relevant sentences from the context that support to answer the question. Then, integrate these sentences to form an answer, ensure the answer is directly extracted from the original text. Output both the identified sentences and the final answer in JSON format. Use \textless{}S\textgreater ~to separate different supporting sentences. Response with the JSON only!  \\
Here is an example: \\
\# Context\\ 
\#\#context\#\#\\ 
\# Question\\ 
Were Scott Derrickson and Ed Wood of the same nationality?\\ 
\# Output\\ 
\{ "supporting\_sentence": "Ed Wood is a 1994 American biographical period comedy-drama film directed and produced by Tim Burton, and starring  Johnny Depp as cult filmmaker Ed Wood. \textless{}S\textgreater{} ~Scott Derrickson (born July 16, 1966) is an American director, screenwriter and producer.",\\ "answer": "yes"\}\\\\
\# Context\\ \#\#context\#\#\\ \# Question\\ \#\#question\#\#\\ \# Output} \\
\hline

MCQA & \makecell[X]{Given a multiple-choice question, select the correct option based on the provided context. First, identify relevant sentences from the context that support to answering the question. Then, integrate these sentences to determine the correct option. Output both the identified supporting sentences and the final selected option in JSON format. Use \textless{}S\textgreater ~to separate different supporting sentences. Response with the JSON only!  \\
Here is an example: \\
\# Context\\ 
\#\#context\#\#
\\\# Question\\ Some people in the Australian outback can't get to a doctor quickly. Because \_\\ \# Options\\ {[}"there are few doctors there", "the nearest doctor is sometimes very far away from them", "there is always heavy traffic on the road", "they don't want to see a doctor"{]}\\ \# Output\\ \{ "supporting\_sentence": "But people in the Australian outback can't get to a doctor quickly.\textless{}S\textgreater{}~The nearest doctor is sometimes hundreds of kilometers away so they have to call him on a two-way radio.",\\ "answer": "the nearest doctor is sometimes very far away from them" \}\\\\
\# Context\\ \#\#context\#\#\\ \# Question\\ \#\#question\#\#\\ \# Options\\ \#\#options\#\#\\ \# Output} \\
\bottomrule
\end{tabularx}
\caption{Prompts for inferring skill evaluation.}
\label{locating-prompt}
\end{table*}

\clearpage

\subsection{Connecting}
\begin{table*}[h!]
\small
\setlength{\arrayrulewidth}{1pt} 
\begin{tabularx}{\textwidth}{X}
\toprule
\textbf{Prompt}\\
\hline
Complete the following passage by selecting and inserting sentences from the provided candidates into the designated blanks, ensuring the passage is coherent and logically structured. The blanks within the passage are denoted by \textless{}NUM\textgreater{}. Output the chosen sentences in the sequence corresponding to the blanks in the passage in JSON format. Response with the JSON only!\\
Here is an example:\\ 
\#passage: \\
Everyone knows Friday the 13th is considered an unlucky day. But why does it have such a bad reputation ? One reason is that both Friday and the number 13 have some troubled ties to Christianity . \textless{}1\textgreater Ever since , the day has been connected with `` bad luck '' . In the Middle Ages , for instance , weddings were not held on Fridays . \textless{}2\textgreater Friday was also unlucky in medieval times because it was `` hangman 's day '' . As for the number 13 , seating 13 people at a table was seen as bad luck because Judas Iscariot , the disciple who betrayed Jesus , is said to have been the 13th guest at The Last Supper . It 's unclear exactly when Friday and the number 13 became connected . \textless{}3\textgreater In 1907 , Thomas Lawson wrote a book titled Friday , the Thirteenth, which described a stockbroker choosing this day to bring down Wall Street. Vyse , who specializes in the psychology of superstitions  , says this is a kind of belief . \textless{}4\textgreater Although there 's evidence that believing in lucky symbols is helpful , Friday the 13th represents a kind of fear . In fact , fear of Friday the 13th has a name : paraskavedekatriaphobia . In Vyse 's view , '' \textless{}5\textgreater\\ 
\#candidates:\\ 
{[}'We only know there are no mentions of Friday the 13th before the 19th century .', \\
'In fact , in Italy,13 is generally considered a lucky number .',\\
'It 's a way for people to ``control the uncontrollable'' and manage the anxiety that comes with uncertain situations.',\\ 
'It was on a Friday that Jesus was put to death .', \\
'Fridays are regarded as an unlucky day and thirteen as an unlucky number.', \\
'We would be better off if no one had ever taught them to us.', \\
'Likewise , it was not a day someone would set out on a journey.'{]}\\ 
\#Output:\\ 
\{\\"output": {[}\\         
\{"position": 1,\\ "sentence": "It was on a Friday that Jesus was put to death ."\},\\         
\{"position": 2,\\ "sentence": "Likewise , it was not a day someone would set out on a journey ."\},\\         
\{"position": 3,\\ "sentence": "We only know there are no mentions of Friday the 13th before the 19th century ."\},\\         
\{"position": 4,\\ "sentence": "It 's a way for people to `` control the uncontrollable '' and manage the anxiety that comes   with uncertain situations ."\},\\        
\{"position": 5,\\ "sentence": "We would be better off if no one had ever taught them to us ."\}{]}\\
\}\\ \\ 
\#passage:\\ \#\#passage\#\#\\ \#candidates:\\ \#\#candidates\#\#\\ \#output:\\
\bottomrule
\end{tabularx}
\caption{Prompt for connecting skill evaluation.}
\label{scde}
\end{table*}
\clearpage

\subsection{Organizing}
\begin{table*}[h!]
\small
\setlength{\arrayrulewidth}{1pt} 
\begin{tabularx}{\textwidth}{X}
\toprule
\textbf{Prompt}\\
\hline
Given a passage and a list of subheadings derived from that passage, your task is to divide the passage into sections, ensuring that each section corresponds to a subheading in the list. You should insert each subheading into the appropriate position in the passage to achieve text segmentation. Output the subheading and its position in turn in JSON format. The key of the JSON should be "output" and its corresponding value should be a list. Response with the JSON only!\\
Here is an example:\\ 
\#Passage: \\
(1) Based on current temperature forecasts, a record-breaking148 million Americansare expected to experience CSI values of 3 or higher on August 2nd. (2) That means their local temperatures would be made at least 3 times more likely because of climate change. (3) Hazardous heat conditions are projected from theRocky Mountainsto thesoutheastern and Mid-Atlantic United States, with late-week and weekend high temperatures reaching themid-to-upper 90sandlow 100s (°F). (4) The National Weather Service forecasts several days of Major to Extreme Heat Risk, from theCentral Plainsto thesoutheastern United States, increasing the likelihood of health impacts for individuals without proper hydration and cooling. (5) High humidity combined with excessive temperatures will lead to dangerous heat index values. (6) Feels-like conditions exceeding110°Fare forecasted across the GreatPlains, Mississippi Valley,andsoutheastern United States. (7) The significant heat dome responsible for these high temperatures is expected to expand across a majority of the Lower 48 statesthrough the first week of August.\\ 
\#Subheadings:\\ 
{[}'Climate Shift Index exposure approaches record', \\
'How unusual is the forecasted heat?'{]}\\ 
\#Output:\\ 
\{"Output": {[}\\         
\{"subheading": 'Climate Shift Index exposure approaches record', "position": 1\},\\         
\{"subheading": 'How unusual is the forecasted heat?', "position": 3\}{]}   \}\\ \\ 
\#Passage:\\ \#\#passage\#\#\\ \#Subheadings:\\ \#\#subheadings\#\#\\ \#Output:\\
\bottomrule
\end{tabularx}
\caption{Prompt for organizing skill evaluation.}
\label{scde}
\end{table*}

\subsection{Selecting}
\begin{table*}[!h]
\small
\setlength{\arrayrulewidth}{1pt} 
\begin{tabularx}{\textwidth}{X}
\toprule
\textbf{Prompt}\\
\hline
Given a passage, you need to select the most important sentences that can serve as a summary of the passage. Sentences in the passage are separated by \textless{}S\textgreater{}. Output these selected sentences in a list format, organized in descending order of importance. Response with the List only!\\
Here is an example:\\ 
\#Passage: \\
13 March 2012 Last updated at 18:31 GMT\textless{}S\textgreater{}Nan Weidong and Nan Weiping have been transforming vegetables into musical instruments for two years.\textless{}S\textgreater{}Their dad was a music teacher and encouraged them to be musical from a young age - but carrot panpipes probably weren't what he had in mind!\textless{}S\textgreater{}Weidong says it's important the veg is fresh - otherwise it risks being out of tune.\textless{}S\textgreater{}And no vegetable is too much of a challenge: they've turned a sweet potato into an ocarina, a bamboo shoot has become a flute, and a yam has doubled up as a whistle.\textless{}S\textgreater{}Watch the clip to see them in action!\\ 
\#Output:\\ 
\{"Output": {[}\\         
"Nan Weidong and Nan Weiping have been transforming vegetables into musical instruments for two years.", 
"Their dad was a music teacher and encouraged them to be musical from a young age - but carrot panpipes probably weren't what he had in mind!", 
"And no vegetable is too much of a challenge: they've turned a sweet potato into an ocarina,  a bamboo shoot has become a flute, and a yam has doubled up as a whistle."{]}
\}\\ \\
\#Passage:\\ \#\#passage\#\#\\ \#Output:\\
\bottomrule
\end{tabularx}
\caption{Prompt for selecting skill evaluation.}
\label{scde}
\end{table*}

\end{document}